\documentclass[10pt,twocolumn,letterpaper]{article}

\usepackage{cvpr}
\usepackage{times}
\usepackage{epsfig}
\usepackage{graphicx}
\usepackage{amsmath}
\usepackage{amssymb}
\usepackage{tabularx}
\usepackage{multirow}

\usepackage[bookmarks=false]{hyperref}
\usepackage{bm}
\usepackage[ labelsep= period]{caption}
\usepackage{booktabs}
\usepackage{subfigure}
\usepackage{enumitem}
\usepackage{xcolor}

\usepackage{lineno,hyperref}
\usepackage{amsmath}
\usepackage{array}
\usepackage{multirow}
\usepackage{algorithm}
\usepackage{algorithmicx}
\usepackage{algpseudocode}
\usepackage{times}
\usepackage{booktabs}
\usepackage{acronym}
\usepackage{threeparttable}
\usepackage{caption}
\usepackage{colortbl}
\usepackage{color}

\usepackage{rotating}

\usepackage{epsfig}
\usepackage{bm}
\usepackage{graphicx}
\usepackage{subfigure}
\usepackage{enumitem}
\usepackage{xcolor}
\usepackage{multirow}
\usepackage{caption}
\def \bbb{\color{blue}}
\def \rrr{\color{red}}
\def \ggg{\color{green}}
\def \kkk{\color{black}}
\definecolor{Gray}{gray}{0.9}
\usepackage{xspace}

\newcommand{\tabincell}[2]{\begin{tabular}{@{}#1@{}}#2\end{tabular}}

\cvprfinalcopy 

\def\cvprPaperID{****} 

\begin{document}

\title{U$^2$-Net: Going Deeper with Nested U-Structure for Salient Object Detection}

\author{Xuebin Qin, Zichen Zhang, Chenyang Huang, Masood Dehghan, Osmar R. Zaiane and Martin Jagersand\\
University of Alberta, Canada\\
{\tt\small \{xuebin,vincent.zhang,chuang8,masood1,zaiane,mj7\}@ualberta.ca}
}

\maketitle

\begin{abstract}
   In this paper, we design a simple yet powerful deep network architecture, U$^2$-Net, for salient object detection (SOD). 
    The architecture of our U$^2$-Net is a two-level nested U-structure. 
    The design has the following advantages: (1) it is able to capture more contextual information from different scales thanks to the mixture of receptive fields of different sizes in our proposed ReSidual U-blocks (RSU), (2) it increases the depth of the whole architecture without significantly increasing the computational cost because of the pooling operations used in these RSU blocks. 
    This architecture enables us to train a deep network from scratch without using backbones from image classification tasks. 
    We instantiate two models of the proposed architecture, U$^2$-Net (176.3 MB, 30 FPS on GTX 1080Ti GPU) and U$^2$-Net$^{\dagger}$ (4.7 MB, 40 FPS), to facilitate the usage in different environments. 
    Both models achieve competitive performance on six SOD datasets. The code is available:\url{https://github.com/NathanUA/U-2-Net}.
\end{abstract}

\section{Introduction}
\label{sec:intro}

Salient Object Detection (SOD) aims at segmenting the most visually attractive objects in an image. 
It is widely used in many fields, such as visual tracking and image segmentation. 
Recently, with the development of deep convolutional neural networks (CNNs), especially the rise of Fully Convolutional Networks (FCN) \cite{long2015fully} in image segmentation, the salient object detection has been improved significantly. 
It is natural to ask, what is still missing?
Let's take a step back and look at the remaining challenges.

There is a common pattern in the design of most SOD networks  \cite{li2016visual,luo2017non,wang2018detect,deng2018r3net}, that is, they focus on making good use of deep features extracted by existing backbones, such as Alexnet~\cite{krizhevsky2012imagenet}, VGG \cite{simonyan2014very}, ResNet \cite{he2016deep}, ResNeXt \cite{xie2017aggregated}, DenseNet \cite{huang2017densely}, etc. 
However, these backbones are all originally designed for image classification. They extract features that are representative of semantic meaning rather than local details and global contrast information, which are essential to saliency detection.
And they need to be pre-trained on ImageNet \cite{imagenet_cvpr09} data which is data-inefficient especially if the target data follows a different distribution than ImageNet.

This leads to our first question: 
\textbf{can we design a new network for SOD, that allows training from scratch and achieves comparable or better performance than those based on existing pre-trained backbones?}

\begin{figure}[t]
    \centering
    \includegraphics[trim=2mm 3mm 2mm 2mm, clip=true, width=0.95\linewidth]{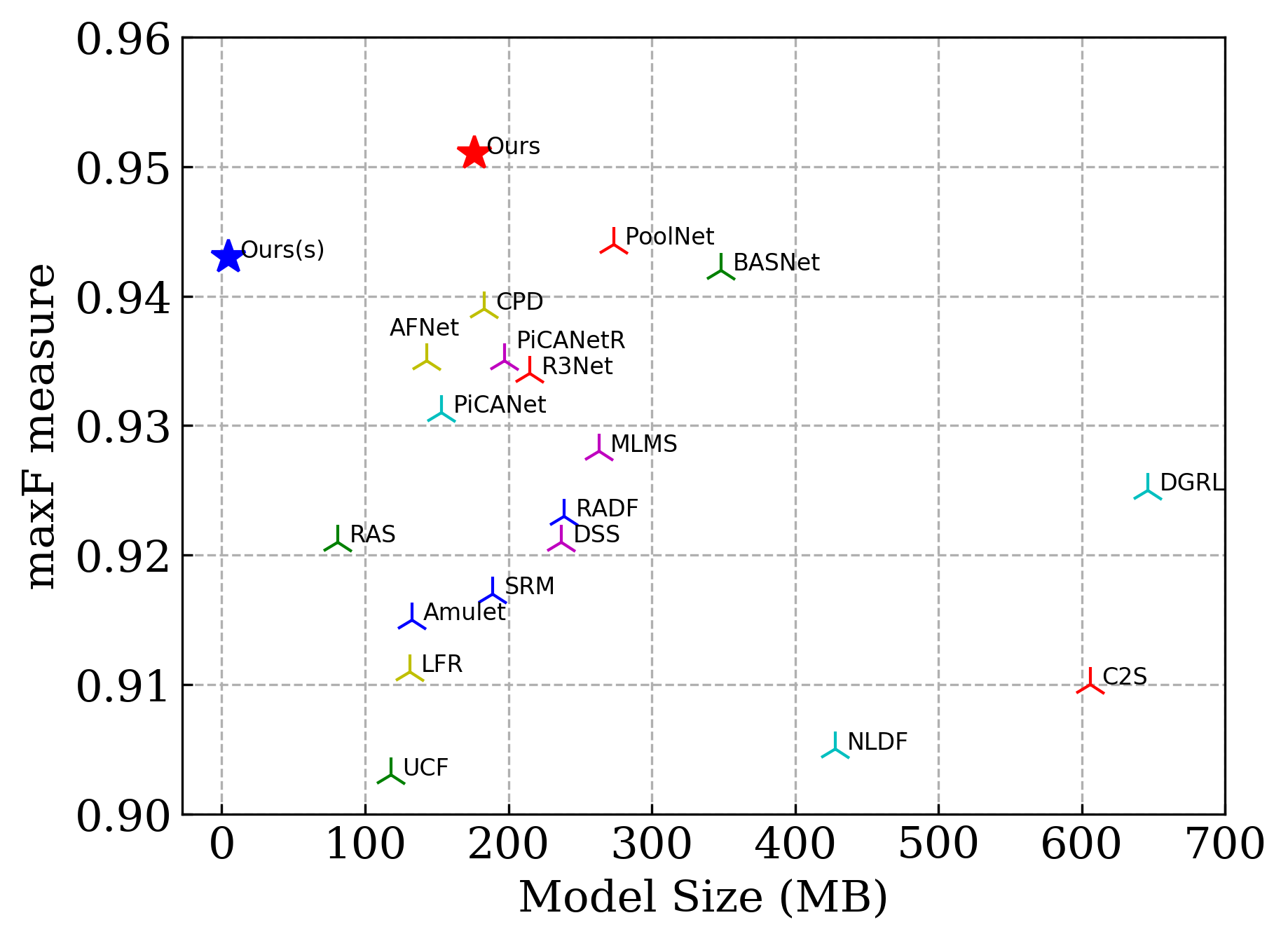}
    \caption{Comparison of model size and performance of our U$^2$-Net with other state-of-the-art SOD models. The $maxF_{\beta}$ measure is computed on dataset ECSSD \cite{yan2013hierarchical}. The red star denotes our U$^2$-Net (Ours) (176.3 MB) and the blue star denotes our small version U$^2$-Net$^{\dagger}$ (Ours$^{\dagger}$) (4.7 MB).}
    \label{fig:sp}
\end{figure}

There are a few more issues on the network architectures for SOD. 
First, they are often overly complicated \cite{zhuge2019deep}. 
It is partially due to the additional feature aggregation modules that are added to the existing backbones to extract multi-level saliency features from these backbones. 
Secondly, the existing backbones usually achieve deeper architecture by sacrificing high resolution of feature maps \cite{zhuge2019deep}. 
To run these deep models with affordable memory and computational cost, the feature maps are down scaled to lower resolution at early stages. 
For instance, at the early layers of both ResNet and DenseNet \cite{huang2017densely}, a convolution with stride of two followed by a maxpooling with stride of two are utilized to reduce the size of the feature maps to one fourth of the input maps. 
However, high resolution also plays an important role in segmentation besides the deep architecture \cite{liu2019auto}.

Hence, our follow-up question is: 
\textbf{can we go deeper while maintaining high resolution feature maps, at a low memory and computation cost?}

Our main contribution is a novel and simple network architecture, called \textbf{U$^2$-Net}, that addresses the two questions above. 
First, U$^2$-Net is a two-level nested U-structure that is designed for SOD without using any pre-trained backbones from image classification. It can be trained from scratch to achieve competitive performance. 
Second, the novel architecture allows the network to go deeper, attain high resolution, without significantly increasing the memory and computation cost. 
This is achieved by a nested U-structure: 
on the bottom level, we design a novel ReSidual U-block (RSU), which is able to extract intra-stage multi-scale features without degrading the feature map resolution; 
on the top level, there is a U-Net like structure, in which each stage is filled by a RSU block. 
The two-level configuration results in a nested U-structure (see Fig. \ref{fig:arc}). 
Our U$^2$-Net (176.3 MB) achieves competitive performance against the state-of-the-art (SOTA) methods on six public datasets, and runs at real-time (30 FPS, with input size of 320$\times$320$\times$3) on a 1080Ti GPU. 
To facilitate the usage of our design in computation and memory constrained environments,  we  provide a small version of our U$^2$-Net, called \textbf{U$^2$-Net$^{\dagger}$} (4.7 MB). 
The U$^2$-Net$^{\dagger}$ achieves competitive results against most of the SOTA models (see Fig. \ref{fig:sp}) at 40 FPS. 

\section{Related Works}
\label{sec:relw}

In recent years, many deep salient object detection networks \cite{PoolNet,BASNet} have been proposed. 
Compared with traditional methods \cite{DBLP:journals/tip/BorjiCJL15} 
based on hand-crafted features like foreground consistency \cite{DBLP:journals/pr/ZhangEWZY17}, hyperspectral information \cite{DBLP:journals/pr/LiangZTBW18}, superpixels' similarity \cite{DBLP:journals/pr/ZhangHLPSH19}, histograms \cite{lu2013robust,lu2012saliency} and so on, deep salient object detection networks show more competitive performance. 

\textbf{Multi-level deep feature integration:} 
Recent works \cite{long2015fully,xie2015holistically} have shown that features from multiple deep layers are able to generate better results \cite{zhang2018bi}. Then, many strategies and methods for integrating and aggregating multi-level deep features are developed for SOD. 
Li \etal (MDF) \cite{li2016visual} propose to feed an image patch around a target pixel to a network and then obtain a feature vector for describing the saliency of this pixel. 
Zhang \etal (Amulet) \cite{amulet17} predict saliency maps by aggregating multi-level features into different resolutions. 
Zhang \etal (UCF) \cite{DBLP:conf/iccv/ZhangWLWY17} propose to reduce the checkerboard artifacts of deconvolution operators by introducing a reformulated dropout and a hybrid upsampling module. 
Luo \etal \cite{luo2017non} design a saliency detection network (NLDF+) with a 4$\times$5 grid architecture, in which deeper features are progressively integrated with shallower features. 
Zhang \etal (LFR) \cite{DBLP:conf/ijcai/ZhangLLS18} predict saliency maps by extracting features from both original input images and their reflection images with a sibling architecture. 
Hou \etal (DSS+) \cite{hou2017deeply} propose to integrate multi-level features by introducing short connections from deep layers to shallow layers. 
Chen \etal (RAS) \cite{DBLP:conf/eccv/ChenTWH18} predict and refine saliency maps by iteratively using the side output saliency of a backbone network as the feature attention guidance. 
Zhang \etal (BMPM) \cite{zhang2018bi} propose to integrate features from shallow and deep layers by a controlled bi-directional passing strategy. 
Deng \etal (R$^3$Net+) \cite{deng2018r3net} alternately incorporate shallow and deep layers' features to refine the predicted saliency maps. 
Hu \etal (RADF+) \cite{DBLP:conf/aaai/HuZQFH18} propose to detect salient objects by recurrently aggregating multi-level deep features. 
Wu \etal (MLMS) \cite{MLMS} improve the saliency detection accuracy by developing a novel Mutual Learning Module for better leveraging the correlation of boundaries and regions. 
Wu \etal \cite{CPD} propose to use Cascaded Partial Decoder (CPD) framework for fast and accurate salient object detection. 
Deep methods in this category take advantage of the multi-level deep features extracted by backbone networks and greatly raise the bar of salient object detection against traditional methods.

\begin{figure*}[t]
    \begin{center}
        \includegraphics[width=1\linewidth]{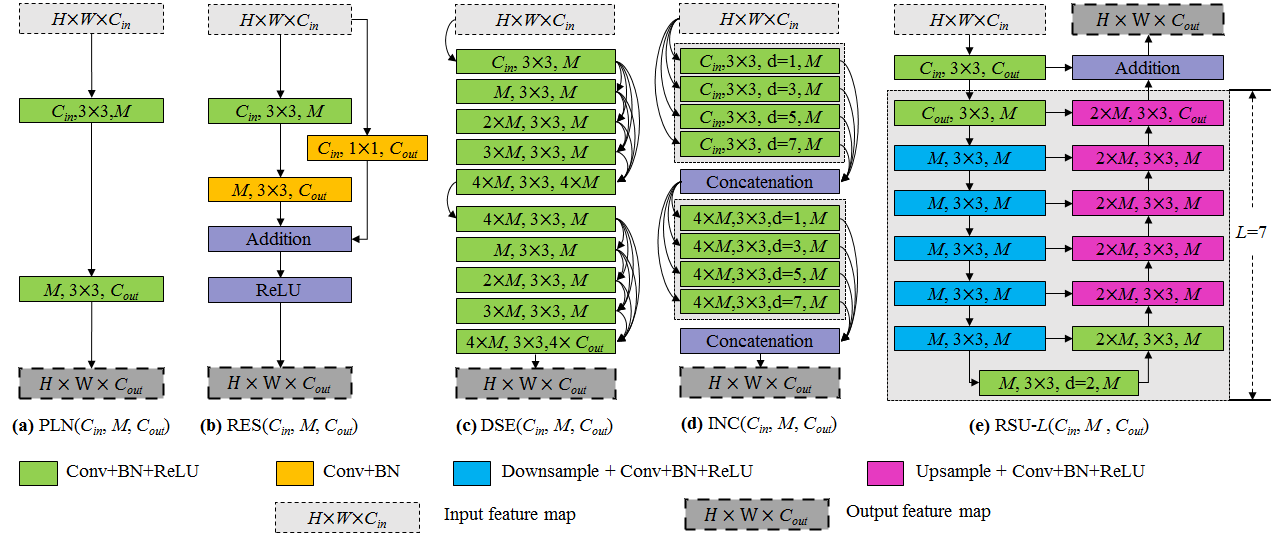}
    \end{center}
    \caption{Illustration of existing convolution blocks and our proposed residual U-block RSU: 
    (a) Plain convolution block PLN, 
    (b) Residual-like block RES, 
    (c) Dense-like block DSE,
    (d) Inception-like block INC and 
    (e) Our residual U-block RSU.
    }
    \label{fig:blks}
\end{figure*}

\textbf{Multi-scale feature extraction: }As mentioned earlier, saliency detection requires both local and global information. 
A $3\times 3$ filter is good for extracting local features at each layer. 
However, it is difficult to extract global information by simply enlarging the filter size because it will increase the number of parameters and computation costs dramatically. 
Many works pay more attention to extracting global context. 
Wang \etal (SRM) \cite{DBLP:conf/iccv/WangBZZL17} adapt the pyramid pooling module \cite{PSPNet} to capture global context and propose a multi-stage refinement mechanism for saliency maps refinement. 
Zhang \etal (PAGRN) \cite{zhang2018progressive} develop a spatial and a channel-wise attention module to obtain the global information of each layer and propose a progressive attention guidance mechanism to refine the saliency maps. 
Wang \etal (DGRL) \cite{wang2018detect} develop an inception-like \cite{szegedy2015going} 
contextual weighting module to localize salient objects globally and then use a boundary refinement module to refine the saliency map locally. 
Liu \etal (PiCANet) \cite{liu2018picanet} recurrently capture the local and global pixel-wise contextual attention and predict the saliency map by incorporating it with a U-Net architecture. 
Zhang \etal (CapSal) \cite{CapSal} design a local and global perception module to extract both local and global information from features extracted by backbone network. 
Zeng \etal (MSWS) \cite{MSWS} design an attention module to predict the spatial distribution of foreground objects over image regions meanwhile aggregate their features.  
Feng \etal (AFNet) \cite{AFNet} develop a global perception module and attentive feedback modules to better explore the structure of salient objects. 
Qin \etal (BASNet) \cite{BASNet} propose a predict-refine model by stacking two differently configured U-Nets sequentially and a Hybrid loss for boundary-aware salient object detection.  
Liu \etal (PoolNet) \cite{PoolNet} develop encoder-decoder architecture for salient object detection by introducing a global guidance module for extraction of global localization features and a multi-scale feature aggregation module adapted from pyramid pooling module for fusing global and fine-level features. In these methods, many inspiring modules are proposed to extract multi-scale features from multi-level deep features extracted from existing backbones. Diversified receptive fields and richer multi-scale contextual features introduced by these novel modules significantly improve the performance of salient object detection models.

In summary, \textbf{multi-level deep feature integration} methods mainly focus on developing better multi-level feature aggregation strategies. On the other hand,  methods in the category of \textbf{multi-scale feature extraction} target at designing new modules for extracting both local and global information from features obtained by backbone networks. As we can see, almost all of the aforementioned methods try to make better use of feature maps generated by the existing image classification backbones. 
Instead of developing and adding more complicated modules and strategies to use these backbones' features, we propose a novel and simple architecture, which directly extracts multi-scale features stage by stage, for salient object detection.

\section{Proposed Method}

First, we introduce the design of our proposed residual U-block and then describe the details of the nested U-architecture built with this block. The network supervision strategy and the training loss are described at the end of this section.

\subsection{Residual U-blocks}
\label{subsec:RSU}

Both local and global contextual information are very important for salient object detection and other segmentation tasks. 
In modern CNN designs, such as VGG, ResNet, DenseNet and so on, small convolutional filters with size of 1$\times$1 or 3$\times$3 are the most frequently used components for feature extraction. They are in favor since they require less storage space and are computationally efficient. 
Figures \ref{fig:blks}(a)-(c) illustrates typical existing convolution blocks with small receptive fields. 
The output feature maps of shallow layers only contain local features because the receptive field of 1$\times$1 or 3$\times$3 filters are too small to capture global information. 
To achieve more global information at high resolution feature maps from shallow layers, the most direct idea is to enlarge the receptive field. 
Fig.~\ref{fig:blks} (d) shows an inception like block \cite{zhang2018bi}, which tries to extract both local and non-local features by enlarging the receptive fields using dilated convolutions \cite{chen2017deeplab}. 
However, conducting multiple dilated convolutions on the input feature map (especially in the early stage) with original resolution requires too much computation and memory resources. 
To decrease the computation costs, PoolNet \cite{PoolNet} adapt the parallel configuration from pyramid pooling modules (PPM) \cite{PSPNet}, which uses small kernel filters on the downsampled feature maps other than the dilated convolutions on the original size feature maps. 
But fusion of different scale features by direct upsampling and concatenation (or addition) may lead to degradation of high resolution features.

Inspired by U-Net \cite{ronneberger2015u}, we propose a novel \textbf{R}e\textbf{S}idual \textbf{U}-block, \textbf{RSU}, to capture intra-stage multi-scale features. 
The structure of RSU-$L$($C_{in},M,C_{out}$) is shown in  Fig.~\ref{fig:blks}(e), where $L$ is the number of layers in the encoder, $C_{in}$, $C_{out}$ denote input and output channels, and $M$ denotes the number of channels in the internal layers of RSU.
Hence, our RSU mainly consists of three components:
\begin{figure}[t]
    \centering
    \includegraphics[trim=1mm 1mm 1mm 2mm, clip=true, width=1\linewidth]{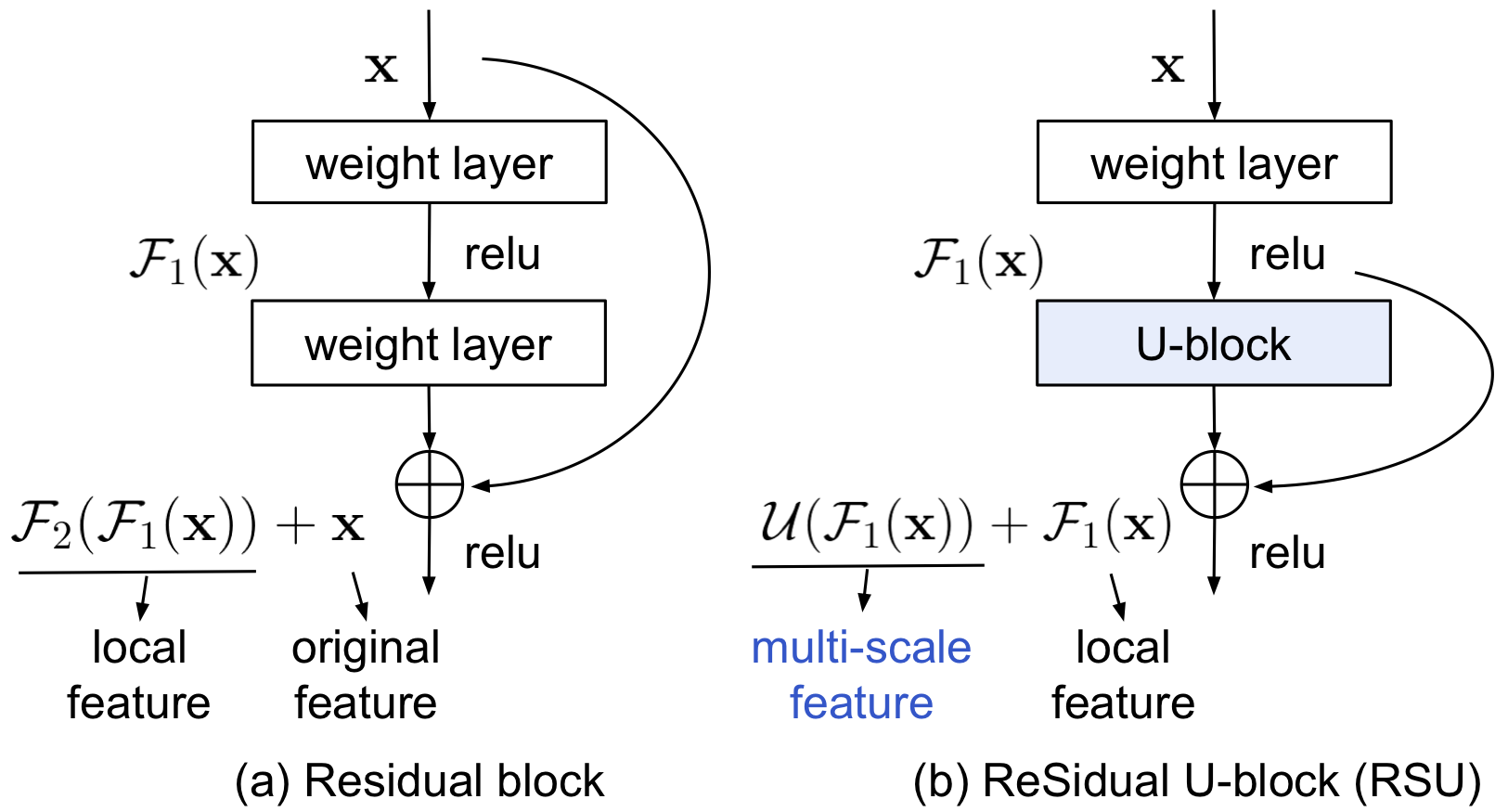} 
    \centering
    \caption{Comparison of the residual block and our RSU.}
    \label{fig:rsu-block}
\end{figure}
\newline \noindent \textbf{(i)} an input convolution layer, which transforms the input feature map $\mathbf{x}$ ($H\times W\times C_{in}$) to an intermediate map $\mathcal{F}_1(\mathbf{x})$ with channel of $C_{out}$. This is a plain convolutional layer for local feature extraction. 
\newline \noindent \textbf{(ii)} a U-Net like symmetric encoder-decoder structure with height of $L$ which takes the intermediate feature map $\mathcal{F}_1(\mathbf{x})$ as input and learns to extract and encode the multi-scale contextual information $\mathcal{U}(\mathcal{F}_1(\mathbf{x}))$. $\mathcal{U}$ represents the U-Net like structure as shown in Fig.~\ref{fig:blks}(e).
Larger $L$ leads to deeper residual U-block (RSU), more pooling operations, larger range of receptive fields and richer local and global features. 
Configuring this parameter enables extraction of multi-scale features from input feature maps with arbitrary spatial resolutions. 
The multi-scale features are extracted from gradually downsampled feature maps and encoded into high resolution feature maps by progressive upsampling, concatenation and convolution.
This process mitigates the loss of fine details caused by direct upsampling with large scales. 
\newline \noindent \textbf{(iii)} a residual connection which fuses local features and the multi-scale features by the summation: $\mathcal{F}_1(\mathbf{x}) + \mathcal{U}(\mathcal{F}_1(\mathbf{x}))$.

\begin{figure}[t]
    \centering
    \includegraphics[width=1\linewidth]{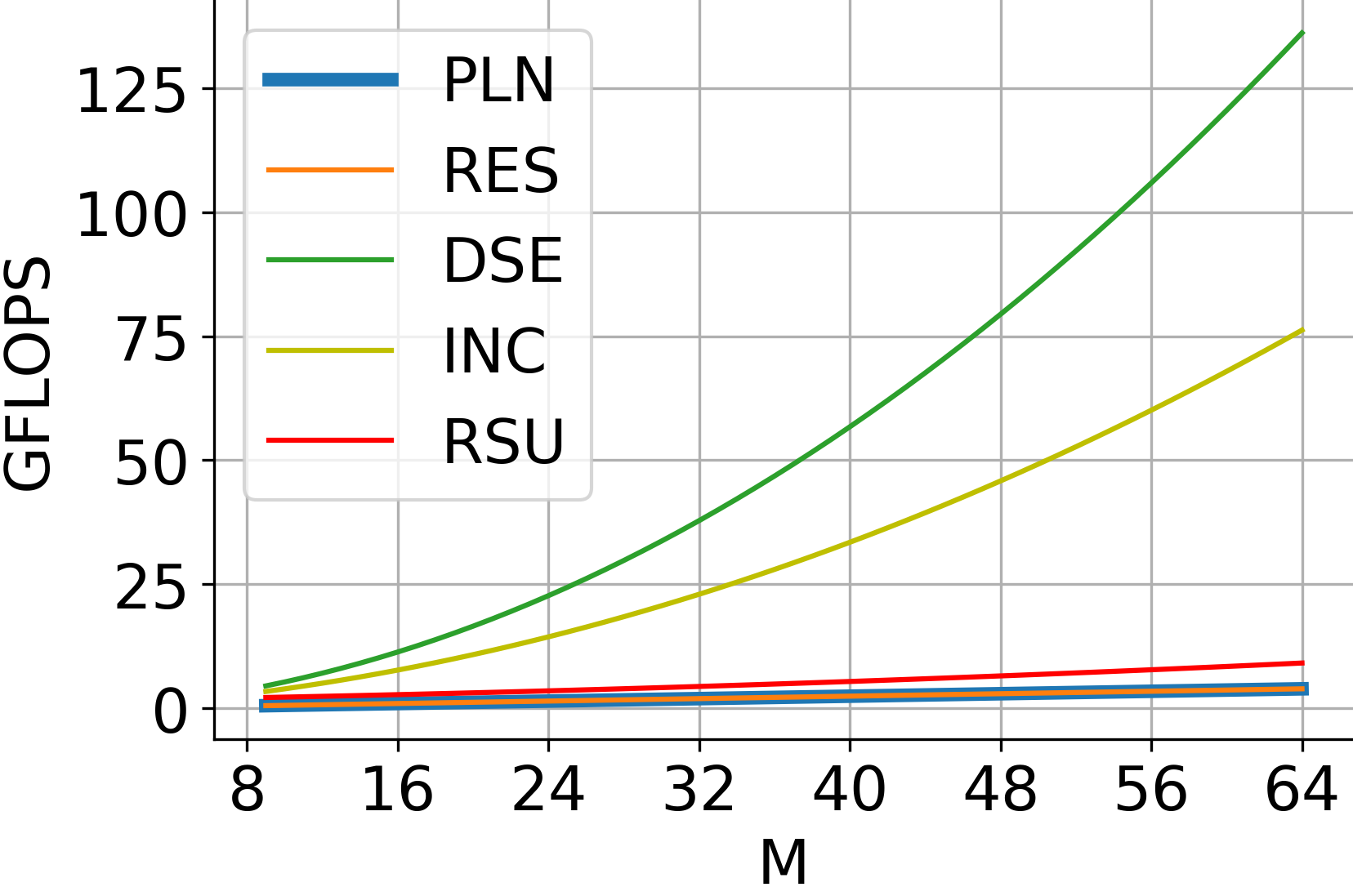} 
    \caption{Computation costs (GFLOPS Giga Floating Point Operations) of different blocks shown in Fig.~\ref{fig:blks}: the computation costs are calculated based on  transferring an input feature map with dimension $320\times 320 \times 3$ to a $320\times 320 \times 64$ output feature map. ``PLN'', ``RES'', ``DSE'', ``INC'' and ``RSU'' denote plain convolution block, residual block, dense block, inception block and our residual U-block respectively.}
    \label{fig:flops}
\end{figure}

To better illustrate the intuition behind our design, we compare our residual U-block (RSU) with the original residual block \cite{he2016deep} in Fig.~\ref{fig:rsu-block}.
The operation in the residual block can be summarized as $\mathcal{H}(\mathbf{x})=\mathcal{F}_2(\mathcal{F}_1(\mathbf{x}))+\mathbf{x}$, where $\mathcal{H}(x)$ denotes the desired mapping of the input features $\mathbf{x}$; $\mathcal{F}_2,\mathcal{F}_1$ stand for the weight layers, which are convolution operations in this setting. 
The main design difference between RSU and residual block is that RSU replaces the plain, single-stream convolution with a U-Net like structure, and replace the original feature with the local feature transformed by a weight layer: 
$\mathcal{H}_{RSU}(\mathbf{x}) = \mathcal{U}(\mathcal{F}_1(\mathbf{x})) + \mathcal{F}_1(\mathbf{x})$, where $\mathcal{U}$ represents the multi-layer U-structure illustrated in Fig.~\ref{fig:blks}(e). This design change empowers the network to extract features from multiple scales directly from  each residual block.
More notably, the computation overhead due to the U-structure is small, since most operations are applied on the downsampled feature maps. 
This is illustrated in Fig.~\ref{fig:flops}, where we show the computation cost comparison between RSU and other feature extraction modules in Fig.~\ref{fig:blks} (a)-(d).
The FLOPs of dense block (DSE), inception block (INC) and RSU all grow quadratically with the number of internal channel $M$. But RSU has a much smaller coefficient on the quadratic term, leading to an improved efficiency. Its computational overhead compared with plain convolution (PLN) and residual block (RES) blocks, which are both linear w.r.t. $M$, is not significant.

\subsection{Architecture of U$^2$-Net}
\label{subsec:arc}

Stacking multiple U-Net-like structures for different tasks has been explored for a while. 
, e.g. stacked hourgalss network \cite{newell2016stacked}, DocUNet \cite{Ma2018DocUNetDI}, CU-Net \cite{tang2018cu} for pose estimation, etc. 
These methods usually stack U-Net-like structures sequentially to build cascaded models and can be summarized as "\textbf{(U$\times n$-Net)}", where $n$ is the number of repeated U-Net modules. The issue is that the computation and the memory costs get magnified by $n$.

\begin{figure*}[h]
    \begin{center}
        \includegraphics[width=0.7\linewidth]{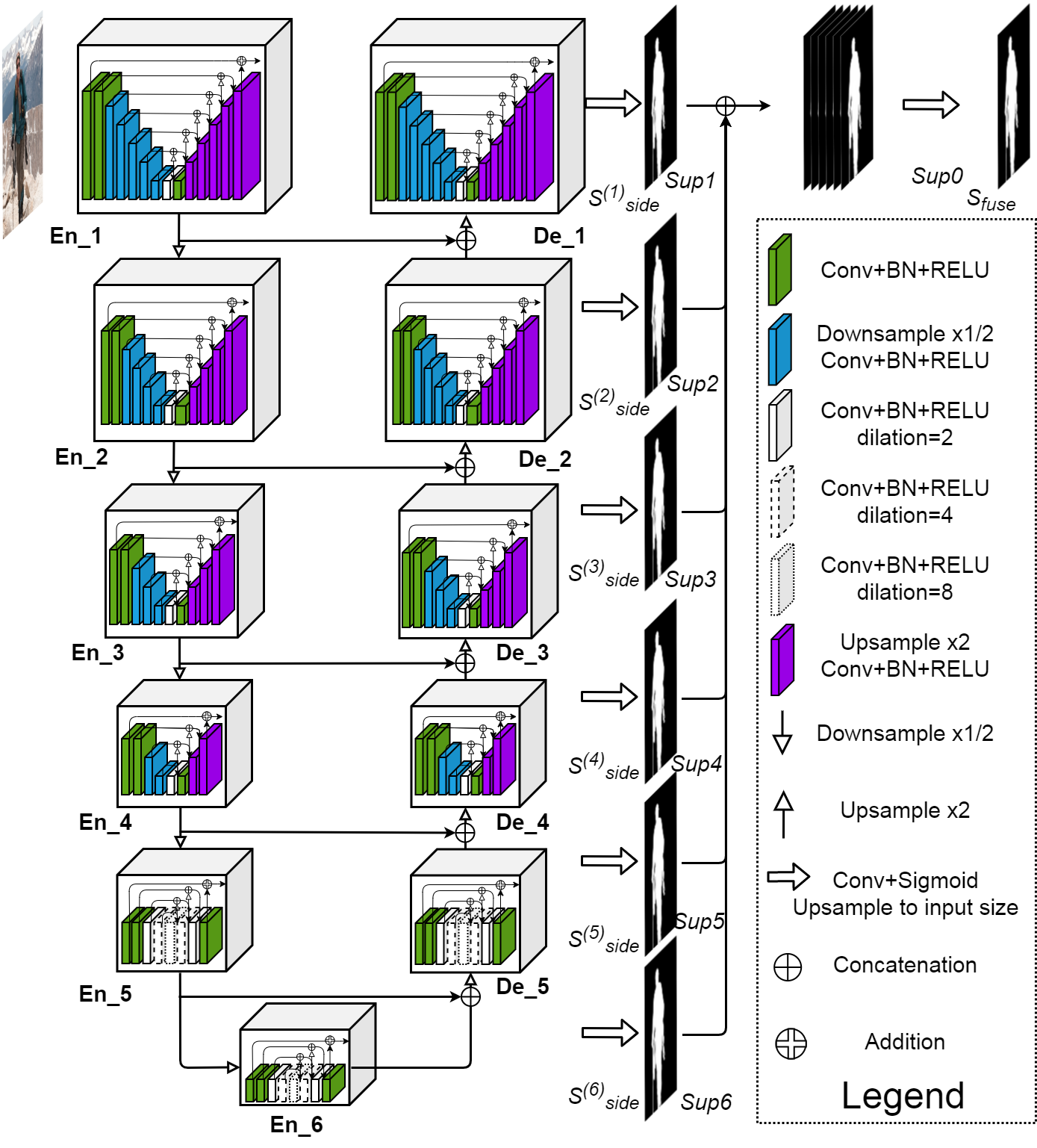}
    \end{center}
    \caption{Illustration of our proposed U$^2$-Net architecture. The main architecture is a U-Net like Encoder-Decoder, where each stage consists of our newly proposed residual U-block (RSU). For example, \textbf{En\_1} is based on our RSU block shown in Fig. \ref{fig:blks}(e). Detailed configuration of RSU block of each stage is given in the last two rows of Table \ref{tab:u2net}.}
    \label{fig:arc}
\end{figure*}

In this paper, we propose a different formulation, \textbf{U$^n$-Net}, of stacking U-structure for salient object detection. 
Our exponential notation refers to nested U-structure rather than cascaded stacking. 
Theoretically, the exponent $n$ can be set as an arbitrary positive integer to achieve single-level or multi-level nested U-structure. 
But architectures with too many nested levels will be too complicated to be implemented and employed in real applications. 

Here, we set $n$ as 2 to build our U$^2$-Net. 
Our U$^2$-Net is a two-level nested U-structure shown in Fig.~\ref{fig:arc}. 
Its top level is a big U-structure consists of 11 stages (cubes in Fig.~\ref{fig:arc}). 
Each stage is filled by a well configured residual U-block (RSU) (bottom level U-structure). 
Hence, the nested U-structure enables the extraction of intra-stage multi-scale features and aggregation of inter-stage multi-level features more efficiently. 

As illustrated in Fig.\ref{fig:arc}, the U$^2$-Net mainly consists of three parts: (1) a six stages encoder, (2) a five stages decoder and (3) a saliency map fusion module attached with the decoder stages and the last encoder stage:

\begin{table*}[tbp] \small
    \centering
    \renewcommand{\arraystretch}{0.9}
    \caption{Detailed configurations of different architectures used in ablation study. ``PLN'', ``RES'', ``DSE'', ``INC'', ``PPM'' and ``RSU'' denote plain convolution block, residual block, dense block, inception block, Pyramid Pooling Module and our residual U-block respectively. ``NIV U$^2$-Net'' denotes U-Net with its each stage replaced by a naive U-Net block. ``I'', ``M'' and ``O'' indicate the number of input channels ($C_{in}$), middle channels and output channels ($C_{out}$) of each block. ``En\_$i$'' and ``De\_$j$'' denote the encoder and decoder stages respectively. The number ``$L$'' in ``NIV-$L$'' and ``RSU-$L$'' denotes the height of the naive U-block and our residual U-block.}
    \resizebox{1\textwidth}{!}{
    \begin{tabular}{cccccccccccc}
        \hline 
        \multirow{2}{*}{\small{\tabincell{c}{\textbf{Architecture with}\\ \textbf{different blocks}}}}& \multicolumn{11}{c}{\textbf{Stages}}\\
        \cline{2-12}
        & \textbf{En\_1} & \textbf{En\_2} & \textbf{En\_3} & \textbf{En\_4} & \textbf{En\_5} & \textbf{En\_6} & \textbf{De\_5} & \textbf{De\_4} & \textbf{De\_3} & \textbf{De\_2} & \textbf{De\_1}\\
        \hline 
        {\small PLN U-Net} & \tabincell{c}{I:3\\M:64\\O:64} & \tabincell{c}{I:64\\M:128\\O:128} & \tabincell{c}{I:128\\M:256\\O:256} & \tabincell{c}{I:256\\M:512\\O:512} & \tabincell{c}{I:512\\M:512\\O:512} & \tabincell{c}{I:512\\M:512\\O:512} & \tabincell{c}{I:1024\\M:512\\O:512} & \tabincell{c}{I:1024\\M:256\\O:256} & \tabincell{c}{I:512\\M:128\\O:128} & \tabincell{c}{I:256\\M:64\\O:64} & \tabincell{c}{I:128\\M:64\\O:64}\\
        \hline
        {\small RES U-Net} & \tabincell{c}{I:3\\M:64\\O:64} & \tabincell{c}{I:64\\M:128\\O:128} & \tabincell{c}{I:128\\M:256\\O:256} & \tabincell{c}{I:256\\M:512\\O:512} & \tabincell{c}{I:512\\M:512\\O:512} & \tabincell{c}{I:512\\M:512\\O:512} & \tabincell{c}{I:1024\\M:512\\O:512} & \tabincell{c}{I:1024\\M:256\\O:256} & \tabincell{c}{I:512\\M:128\\O:128} & \tabincell{c}{I:256\\M:64\\O:64} & \tabincell{c}{I:128\\M:64\\O:64}\\
        \hline
        {\small DSE U-Net} & \tabincell{c}{I:3\\M:32\\O:64} & \tabincell{c}{I:64\\M:32\\O:128} & \tabincell{c}{I:128\\M:64\\O:256} & \tabincell{c}{I:256\\M:128\\O:512} & \tabincell{c}{I:512\\M:128\\O:512} & \tabincell{c}{I:512\\M:128\\O:512} & \tabincell{c}{I:1024\\M:128\\O:512} & \tabincell{c}{I:1024\\M:64\\O:256} & \tabincell{c}{I:512\\M:32\\O:128} & \tabincell{c}{I:256\\M:16\\O:64} & \tabincell{c}{I:128\\M:16\\O:64}\\
        \hline
        {\small INC U-Net} & \tabincell{c}{I:3\\M:32\\O:64} & \tabincell{c}{I:64\\M:32\\O:128} & \tabincell{c}{I:128\\M:64\\O:256} & \tabincell{c}{I:256\\M:128\\O:512} & \tabincell{c}{I:512\\M:128\\O:512} & \tabincell{c}{I:512\\M:128\\O:512} & \tabincell{c}{I:1024\\M:128\\O:512} & \tabincell{c}{I:1024\\M:64\\O:256} & \tabincell{c}{I:512\\M:32\\O:128} & \tabincell{c}{I:256\\M:16\\O:64} & \tabincell{c}{I:128\\M:16\\O:64}\\
        \hline
       {\small PPM U-Net} & \tabincell{c}{I:3\\M:32\\O:64} & \tabincell{c}{I:64\\M:32\\O:128} & \tabincell{c}{I:128\\M:64\\O:256} & \tabincell{c}{I:256\\M:128\\O:512} & \tabincell{c}{I:512\\M:128\\O:512} & \tabincell{c}{I:512\\M:128\\O:512} & \tabincell{c}{I:1024\\M:128\\O:512} & \tabincell{c}{I:1024\\M:64\\O:256} & \tabincell{c}{I:512\\M:32\\O:128} & \tabincell{c}{I:256\\M:16\\O:64} & \tabincell{c}{I:128\\M:16\\O:64} \kkk\\ 
        \hline
        {\small NIV U$^2$-Net} & \tabincell{c}{NIV-7\\I:3\\M:32\\O:64} & \tabincell{c}{NIV-6\\I:64\\M:32\\O:128} & \tabincell{c}{NIV-5\\I:128\\M:64\\O:256} & \tabincell{c}{NIV-4\\I:256\\M:128\\O:512} & \tabincell{c}{NIV-4F\\I:512\\M:256\\O:512} & \tabincell{c}{NIV-4F\\I:512\\M:256\\O:512} & \tabincell{c}{NIV-4F\\I:1024\\M:256\\O:512} & \tabincell{c}{NIV-4\\I:1024\\M:128\\O:256} & \tabincell{c}{NIV-5\\I:512\\M:64\\O:128} & \tabincell{c}{NIV-6\\I:256\\M:32\\O:64} & \tabincell{c}{NIV-7\\I:128\\M:16\\O:64}\\
        \hline
        {\small U$^2$-Net (\textbf{Ours})} & \tabincell{c}{RSU-7\\I:3\\M:32\\O:64} & \tabincell{c}{RSU-6\\I:64\\M:32\\O:128} & \tabincell{c}{RSU-5\\I:128\\M:64\\O:256} & \tabincell{c}{RSU-4\\I:256\\M:128\\O:512} & \tabincell{c}{RSU-4F\\I:512\\M:256\\O:512)} & \tabincell{c}{RSU-4F\\I:512\\M:256\\O:512)} & \tabincell{c}{RSU-4F\\I:1024\\M:256\\O:512} & \tabincell{c}{RSU-4\\I:1024\\M:128\\O:256} & \tabincell{c}{RSU-5\\I:512\\M:64\\O:128} & \tabincell{c}{RSU-6\\I:256\\M:32\\O:64} &  \tabincell{c}{RSU-7\\I:128\\M:16\\O:64}\\
        \hline
        {\small U$^2$-Net$^{\dagger}$ (\textbf{Ours$^{\dagger}$})} & \tabincell{c}{RSU-7\\I:3\\M:16\\O:64} & \tabincell{c}{RSU-6\\I:64\\M:16\\O:64} & \tabincell{c}{RSU-5\\I:64\\M:16\\O:64} & \tabincell{c}{RSU-4\\I:64\\M:16\\O:64} & \tabincell{c}{RSU-4F\\I:64\\M:16\\O:64} & \tabincell{c}{RSU-4F\\I:64\\M:16\\O:64} & \tabincell{c}{RSU-4F\\I:128\\M:16\\O:64} & \tabincell{c}{RSU-4\\I:128\\M:16\\O:64} & \tabincell{c}{RSU-5\\I:128\\M:16\\O:64} & \tabincell{c}{RSU-6\\I:128\\M:16\\O:64} &  \tabincell{c}{RSU-7\\I:128\\M:16\\O:64}\\
        \hline
    \end{tabular}
    }
    \label{tab:u2net}
\end{table*}

\noindent 
(i) In encoder stages \textbf{En\_1}, \textbf{En\_2}, \textbf{En\_3} and \textbf{En\_4}, we use residual U-blocks RSU-7, RSU-6, RSU-5 and RSU-4, respectively. As mentioned before, ``7'', ``6'', ``5'' and ``4'' denote the heights ($L$) of RSU blocks. The $L$ is usually configured according to the spatial resolution of the input feature maps. 
For feature maps with large height and width, we use greater $L$ to capture more large scale information. The resolution of feature maps in \textbf{En\_5} and \textbf{En\_6} are relatively low, further downsampling of these feature maps leads to loss of useful context. Hence, in both \textbf{En\_5} and \textbf{En\_6} stages, RSU-4F are used, where ``F'' means that the RSU is a dilated version, in which we replace the pooling and upsampling operations with dilated convolutions (see Fig. \ref{fig:arc}). That means all of intermediate feature maps of RSU-4F have the same resolution with its input feature maps. 

\noindent
(ii) The decoder stages have similar structures to their symmetrical encoder stages with respect to \textbf{En\_6}. In \textbf{De\_5}, we also use the dilated version residual U-block RSU-4F which is similar to that used in the encoder stages \textbf{En\_5} and \textbf{En\_6}. Each decoder stage takes the concatenation of the upsampled feature maps from its previous stage and those from its symmetrical encoder stage as the input, see Fig.~\ref{fig:arc}. 
 
 \noindent
 (iii) The last part is the saliency map fusion module which is used to generate saliency probability maps. Similar to HED \cite{xie2015holistically}, our U$^2$-Net first generates six side output saliency probability maps $\mathcal{S}_{side}^{(6)}$, $\mathcal{S}_{side}^{(5)}$, $\mathcal{S}_{side}^{(4)}$, $\mathcal{S}_{side}^{(3)}$, $\mathcal{S}_{side}^{(2)}$, $\mathcal{S}_{side}^{(1)}$ from stages \textbf{En\_6}, \textbf{De\_5}, \textbf{De\_4}, \textbf{De\_3}, \textbf{De\_2} and \textbf{De\_1} by a $3\times 3$ convolution layer and a sigmoid function. Then, it upsamples the logits (convolution outputs before sigmoid functions) of the side output saliency maps to the input image size and fuses them with a concatenation operation followed by a $1\times 1$ convolution layer and a sigmoid function to generate the final saliency probability map $\mathcal{S}_{fuse}$ (see bottom right of Fig. \ref{fig:arc}).

In summary, the design of our U$^2$-Net allows having deep architecture with rich multi-scale features and relatively low computation and memory costs. 
In addition, since our U$^2$-Net architecture is only built upon our RSU blocks without using any pre-trained backbones adapted from image classification, it is flexible and easy to be adapted to different working environments with insignificant performance loss. 
In this paper, we provide two instances of our U$^2$-Net by using different configurations of filter numbers: a normal version \textbf{U$^2$-Net} (176.3 MB) and a relatively smaller version \textbf{U$^2$-Net$^{\dagger}$} (4.7 MB). Detailed configurations are presented in the last two rows of Table \ref{tab:u2net}. 

\subsection{Supervision}
\label{subsec:sup}

In the training process, we use deep supervision similar to HED \cite{xie2015holistically}. 
Its effectiveness has been proven in HED and DSS. 
Our training loss is defined as:

\begin{equation}
    \centering 
    \mathcal{L} = \sum_{m=1}^M{w^{(m)}_{side}\ell^{(m)}_{side}} + w_{fuse}\ell_{fuse}
    \label{equ:ls_all}
\end{equation}
\noindent
where $\ell_{side}^{(m)}$ ($M=6$, as the Sup1, Sup2, $\cdots$, Sup6 in Fig. \ref{fig:arc}) is the loss of the side output saliency map $S^{(m)}_{side}$ and $\ell_{fuse}$ (Sup7 in Fig.~\ref{fig:arc}) is the loss of the final fusion output saliency map $S_{fuse}$. $w^{(m)}_{side}$ and $w_{fuse}$ are the weights of each loss term. 
For each term $\ell$, we use the standard binary cross-entropy to calculate the loss:

\begin{equation}
    \small{
    \ell = -\sum_{(r,c)}^{(H,W)}[P_{G(r,c)}logP_{S(r,c)}+(1-P_{G(r,c)})log(1-P_{S(r,c)})]}
    \label{equ:ls_bce}
\end{equation}
\noindent
where $(r,c)$ is the pixel coordinates and $(H,W)$ is image size: height and width. $P_{G(r,c)}$ and $P_{S(r,c)}$ denote the pixel values of the ground truth and the predicted saliency probability map, respectively. 
The training process tries to minimize the overall loss $\mathcal{L}$ of Eq.~(\ref{equ:ls_all}). 
In the testing process, we choose the fusion output $\ell_{fuse}$ as our final saliency map.

\section{Experimental Results}

\subsection{Datasets}

\textbf{Training dataset:} We train our network on \textbf{DUTS-TR}, which is a part of DUTS dataset \cite{wang2017learning}. \textbf{DUTS-TR} contains 10553 images in total. Currently, it is the largest and most frequently used training dataset for salient object detection.
We augment this dataset by horizontal flipping to obtain 21106 training images offline. 

\textbf{Evaluation datasets:} Six frequently used benchmark datasets are used to evaluate our method including: DUT-OMRON \cite{yang2013saliency}, DUTS-TE \cite{wang2017learning}, HKU-IS \cite{li2016visual}, ECSSD \cite{yan2013hierarchical}, PASCAL-S \cite{li2014secrets}, SOD \cite{movahedi2010design}. 
\textbf{DUT-OMRON} includes 5168 images, most of which contain one or two structurally complex foreground objects.
DUTS dataset consists of two parts: DUTS-TR and DUTS-TE. 
As mentioned above we use DUTS-TR for training. Hence, \textbf{DUTS-TE}, which contains 5019 images, is selected as one of our evaluation dataset.
\textbf{HKU-IS} contains 4447 images with multiple foreground objects. 
\textbf{ECSSD} contains 1000 structurally complex images and many of them contain large foreground objects. 
\textbf{PASCAL-S} contains 850 images with complex foreground objects and cluttered background.
\textbf{SOD} only contains 300 images. But it is very challenging. 
Because it was originally designed for image segmentation and many images are low contrast or contain complex foreground objects overlapping with the image boundary. 

\subsection{Evaluation Metrics}

The outputs of the deep salient object methods are usually probability maps that have the same spatial resolution with the input images. 
Each pixel of the predicted saliency maps has a value within the range of 0 and 1 (or [0, 255]).
The ground truth are usually binary masks, in which each pixel is either 0 or 1 (or 0 and 255) where 0 indicates the background pixels and 1 indicates the foreground salient object pixels. 

To comprehensively evaluate the quality of those probability maps against the ground truth, six measures including (1) Precision-Recall (PR) curves
, (2) maximal F-measure ($maxF_{\beta}$) \bbb \cite{freqbased} \kkk, (3) Mean Absolute Error ($MAE$) \cite{liu2018picanet,BASNet,PoolNet}, (4) weighted F-measure ($F^w_\beta$) \bbb \cite{Margolin2014HowTE} \kkk, (5) structure measure ($S_m$) \cite{fan2017structure} and (6) relaxed F-measure of boundary ($\text{\textit{relax}}F^b_\beta$) \cite{BASNet} are used: 

\noindent \textbf{(1) PR curve} is plotted based on a set of precision-recall pairs. 
Given a predicted saliency probability map, its precision and recall scores are computed by comparing its thresholded binary mask against the ground truth mask. 
The precision and recall of a dataset are computed by averaging the precision and recall scores of those saliency maps. 
By varying the thresholds from 0 to 1, we can obtain a set of average precision-recall pairs of the dataset. 

\noindent \textbf{(2) F-measure} $F_\beta$ is used to comprehensively evaluate both precision and recall as:

\begin{equation}
    \centering
    F_\beta = \tfrac{(1+\beta^2)\times Precision\times Recall}{\beta^2\times Precision + Recall}.
    \label{equ:fscore}
\end{equation}
\noindent
We set the $\beta^2$ to 0.3 and report the maximum $F_\beta$ ($\text{\textit{max}}F_\beta$) for each dataset similar to previous works \cite{freqbased,liu2018picanet,zhang2018bi}.

\noindent \textbf{(3) \textit{MAE}} is the Mean Absolute Error which denotes the average per-pixel difference between a predicted saliency map and its ground truth mask. It is defined as:

\begin{equation}
\centering
MAE = \tfrac{1}{H\times W}{\textstyle \sum}_{r=1}^H {\textstyle \sum}_{c=1}^W{|P(r,c)-G(r,c)|}
\label{equ:MAE}
\end{equation}
\noindent
where $P$ and $G$ are the probability map of the salient object detection and the corresponding ground truth respectively, ($H$, $W$) and $(r,c)$ are the (height, width) and the pixel coordinates. 

\noindent \textbf{(4) weighted F-measure} ($F^w_\beta$) \cite{Margolin2014HowTE} 
is utilized as a complementary measure to $\text{\textit{max}}F_\beta$ for overcoming the possible unfair comparison caused by ``interpolation flaw, dependency flaw and equal-importance flaw'' \cite{liu2018picanet}. It is defined as:

\begin{equation}
	\centering
	F^w_\beta = (1+\beta^2)\frac{Precision^w\cdot Recall^w}{\beta^2\cdot Precision^w+Recall^w}.
\end{equation}

\noindent \textbf{(5) S-measure} ($S_m$) is used to evaluate the structure similarity of the predicted non-binary saliency map and the ground truth. 
The S-measure is defined as the weighted sum of region-aware $S_r$ and object-aware $S_o$ structural similarity: 

\begin{equation}
	\centering
	S = (1-\alpha) S_r + \alpha S_o.
\end{equation} 
\noindent
where $\alpha$ is usually set to 0.5.

\noindent \textbf{(6) relax boundary F-measure} $\text{\textit{relax}}F^b_\beta$ \cite{ehrig2005relaxed} 
is utilized to quantitatively evaluate boundaries' quality of the predicted saliency maps \cite{BASNet}. 
Given a saliency probability map $P \in[0,1]$, its binary mask $P_{bw}$ is obtained by a simple thresholding operation (threshold is set to $0.5$). 
Then, the $XOR(P_{bw},P_{erd})$ operation is conducted to obtain its one pixel wide boundary, where $P_{erd}$ denotes the eroded binary mask \cite{haralick1987image} 
of $P_{bw}$. 
The boundaries of ground truth mask are obtained in the same way. 
The computation of relaxed boundary F-measure $\text{\textit{relax}}F^b_\beta$ is similar to equation (\ref{equ:fscore}). The difference is that $\text{\textit{relax}}Precision^b$ and $\text{\textit{relax}}Recall^b$ other than $Precision$ and $Recall$ are used in equation (\ref{equ:fscore}). 
The definition of relaxed boundary precision ($\text{\textit{relax}}Precision^b$) is the fraction of predicted boundary pixels within a range of $\rho$ pixels from ground truth boundary pixels. 
The relaxed boundary recall ($\text{\textit{relax}}Recall^b$) is defined as the fraction of ground truth boundary pixels that are within $\rho$ pixels of predicted boundary pixels. 
The slack parameter $\rho$ is set to 3 as in the previous work \cite{BASNet}. 
Given a dataset, its average $\text{\textit{relax}}F^b_\beta$ of all predicted saliency maps is reported in this paper. 

\subsection{Implementation Details}

In the training process, each image is first resized to 320$\times$320 and randomly flipped vertically and cropped to 288$\times$288. 
We are not using any existing backbones in our network. 
Hence, we train our network from scratch and all of our convolutional layers are initialized by Xavier \cite{DBLP:journals/jmlr/GlorotB10}. 
The loss weights $w_{side}^{(m)}$ and $w_{fuse}$ are all set to 1. 
Adam optimizer \cite{kingma2014adam} 
is used to train our network and its hyper parameters are set to default (initial learning rate lr=1e-3, betas=(0.9, 0.999), eps=1e-8, weight\_decay=0). 
We train the network until the loss converges without using validation set which follows the previous methods \cite{PoolNet,liu2018picanet,zhang2018bi}. 
After 600k iterations (with a batch size of 12), the training loss converges and the whole training process takes about 120 hours. 
During testing, the input images ($H\times W$) are resized to 320$\times$320 and fed into the network to obtain the saliency maps. The predicted saliency maps with size of 320$\times$320 are resized back to the original size of the input image ($H\times W$). 
Bilinear interpolation is used in both resizing processes. 
Our network is implemented based on Pytorch 0.4.0 \cite{paszke2017automatic}. 
Both training and testing are conducted on an eight-core, 16 threads PC with an AMD Ryzen 1800x 3.5 GHz CPU (32GB RAM) and a GTX 1080ti GPU (11GB memory). 
We will release our code later. 

\begin{table}[h] \small
    \centering
    \caption{Results of ablation study on different blocks, architectures and backbones. ``PLN'', ``RES'', ``DSE'', ``INC'', ``PPM'' and ``RSU'' denote plain convolution block, residual block, dense block, inception block, pyramid pooling module and our residual U-block respectively. ``NIV U$^2$-Net'' denotes U-Net with its each stage replaced by a naive U-Net block. The ``Time (ms)'' (ms: milliseconds) costs are computed by averaging the inference time costs of images in ECSSD dataset. Values with bold fonts indicate the best two performance.
    } 
    \resizebox{0.5\textwidth}{!}{
    \begin{tabular}{rccccc}
		\hline
        \multirow{2}{*}{{\small Configuration}} & \multicolumn{2}{c}{DUT-OMRON} & \multicolumn{2}{c}{ECSSD} & \multirow{2}{*}{{\small Time (ms)}}\\
        \cline{2-5}
         & {\small $\text{\textit{max}}F_\beta$} & {\small $MAE$} & {\small $\text{\textit{max}}F_\beta$} & {\small $MAE$}\\
        \hline
        \multicolumn{1}{r}{Baseline U-Net} &  0.725 &  0.082 & 0.896 & 0.066 & \textbf{14} \kkk\\
        \multicolumn{1}{r}{PLN U-Net} &  0.782 &  0.062 & 0.928 & 0.043 & \textbf{16} \kkk\\
        \multicolumn{1}{r}{RES U-Net} &  0.781 & 0.065 & 0.933 & 0.042 & 19 \\
        \multicolumn{1}{r}{DSE U-Net} &  0.790 & 0.067 & 0.927 & 0.046 & 70 \\
        \multicolumn{1}{r}{INC U-Net} &  0.777 & 0.069 & 0.921 & 0.047 & 57 \kkk\\
        \multicolumn{1}{r}{ PPM U-Net} & 0.792 & 0.062 & 0.928 & 0.049 & 105 \\
        \hline 
        \multicolumn{1}{r}{ Stacked HourglassNet \cite{newell2016stacked}} &  0.756 & 0.073 & 0.905 & 0.059 & 103 \\
        \multicolumn{1}{r}{CU-NET \cite{tang2018quantized}} & 0.767 & 0.072 & 0.913 & 0.061 & 50 \\
        \multicolumn{1}{r}{NIV U$^2$-Net} & 0.803 & 0.061 & 0.938 & 0.085 & 30\\
        \hline 
        \multicolumn{1}{r}{U$^2$-Net w/ VGG-16 backbone} & 0.808 & 0.063 & 0.942 & 0.038 & 23 \\
        \multicolumn{1}{r}{U$^2$-Net w/ ResNet-50 backbone} & 0.813 & 0.058 & 0.937 & 0.041 & 41 \\
        \hline 
        \multicolumn{1}{r}{(Ours) RSU U$^2$-Net} &  \textbf{0.823} & \textbf{0.054} & \textbf{0.951} & \textbf{0.033} & 33 \\
        \multicolumn{1}{r}{(Ours$^{\dagger}$) RSU U$^2$-Net$^{\dagger}$} &  \textbf{0.813} & \textbf{0.060} & \textbf{0.943} & \textbf{0.041} & 25\\
        \hline 
    \end{tabular}
    }
    \label{tab:ablation}
\end{table}

\subsection{Ablation Study}
\label{sec:abl}

To verify the effectiveness of our U$^2$-Net, ablation studies are conducted on the following three aspects: i) basic blocks, ii) architectures and iii) backbones. All the ablation studies follow the same implementation setup.

\subsubsection{Ablation on Blocks}
In the blocks ablation, the goal is to validate the effectiveness of our newly designed residual U-blocks (RSUs). Specifically, we fix the outside Encoder-Decoder architecture of our U$^2$-Net and replace its stages with other popular blocks including plain convolution blocks (PLN), residual-like blocks (RSE), dense-like blocks (DSE), inception-like blocks (INC) and pyramid pooling module (PPM) other than RSU block, as shown in Fig.~\ref{fig:blks} (a)-(d). 
Detailed configurations can be found in Table~\ref{tab:u2net}. 

Table~\ref{tab:ablation} shows the quantitative results of the ablation study. 
As we can see, the performance of baseline U-Net is the worst, while PLN U-Net, RES U-Net, DES U-Net, INC U-Net and PPM U-Net achieve better performance than the baseline U-Net. Because they are either deeper or have the capability of extracting multi-scale features. However, their performance is still inferior to both our full size U$^2$-Net and small version U$^2$-Net$^{\dagger}$. Particularly, our full size U$^2$-Net improves the $\text{\textit{max}}F_\beta$ about 3.3\% and 1.8\%, and decreases the $MAE$ over 12.9\% and 21.4\% against the second best model (in the blocks ablation study) on DUT-OMRON and ECSSD datasets, respectively. 
Furthermore, our U$^2$-Net and U$^2$-Net$^{\dagger}$ increase the $\text{\textit{max}}F_\beta$ by 9.8\% and 8.8\% and decrease the $MAE$ by 34.1\% and 27.0\%, which are significant improvements, on DUT-OMRON dataset against the baseline U-Net. 
On ECSSD dataset, although the $\text{\textit{max}}F_\beta$ improvements (5.5\%, 4.7\%) of our U$^2$-Net and U$^2$-Net$^{\dagger}$ against the baseline U-Net is slightly less significant than that on DUT-OMRON, the improvements of $MAE$ are much greater (50.0\%, 38.0\%). 
Therefore, we believe that our newly designed residual U-block RSU is better then others in this salient object detection task. Besides, there is no significant time costs increasing of our residual U-block (RSU) based U$^2$-Net architectures. 
\subsubsection{Ablation on Architectures} 

As we mentioned above, previous methods usually use cascaded ways to stack multiple similar structures for building more expressive models. One of the intuitions behind this idea is that multiple similar structures are able to refine the results gradually while reducing overfitting. Stacked HourglassNet \cite{newell2016stacked} and CU-Net \cite{tang2018quantized} are two representative models in this category. Therefore, we adapted the stacked HourglassNet and CU-Net to compare the performance between the cascaded architectures and our nested architectures. As shown in Table.~\ref{tab:ablation}, both our full size U$^2$-Net and small size model U$^2$-Net$^\dagger$ outperform these two cascaded models. 
It is worth noting the both stacked HourglassNet and CU-Net utilizes improved U-Net-like modules as their stacking sub-models. To further demonstrate the effectiveness of our nested architecture, we also illustrate the performance of an U$^2$-Net based on naive U-blocks (NIV) other than our newly proposed residual U-blocks. We can see that the NIV U$^2$-Net still achieves better performance than these two cascaded models. In addition, the nested architectures are faster than the cascaded ones. In summary, our nested architecture is able to achieve better performance than the cascaded architecture both in terms of accuracy and speed. 

\subsubsection{Ablation on Backbones} 

Different from the previous salient object detection models which use backbones (\emph{e.g.} VGG, ResNet, etc.) as their encoders, our newly proposed U$^2$-Net architecture is backbone free. To validate the backbone free design, we conduct ablation studies on replacing the encoder part of our full size U$^2$-Net with different backbones: VGG16 and ResNet50. Practically, we adapt the backbones (VGG-16 and ResNet-50) by adding an extra stage after their last convolutional stages to achieve the same receptive fields with our original U$^2$-Net architecture design. As shown in Table \ref{tab:ablation}, the models using backbones and our RSUs as decoders achieve better performance than the previous ablations and comparable performance against our small size U$^2$-Net. However, they are still inferior to our full size U$^2$-Net. Therefore, we believe that our backbones free design is more competitive than backbones-based design in this salient object detection task. 

\begin{figure*}
    \centering
    \includegraphics[width=0.49\linewidth]{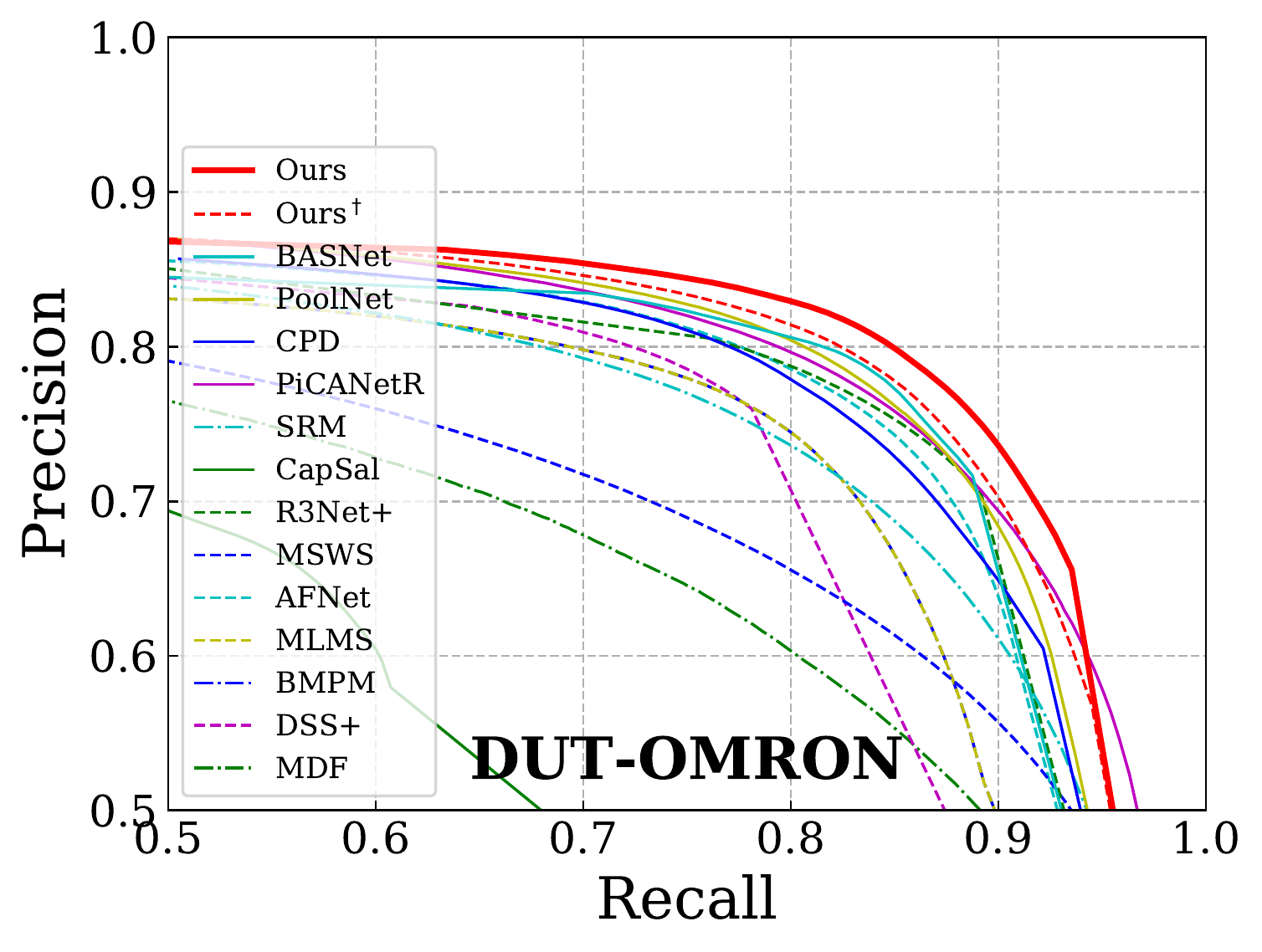}
    \includegraphics[width=0.49\linewidth]{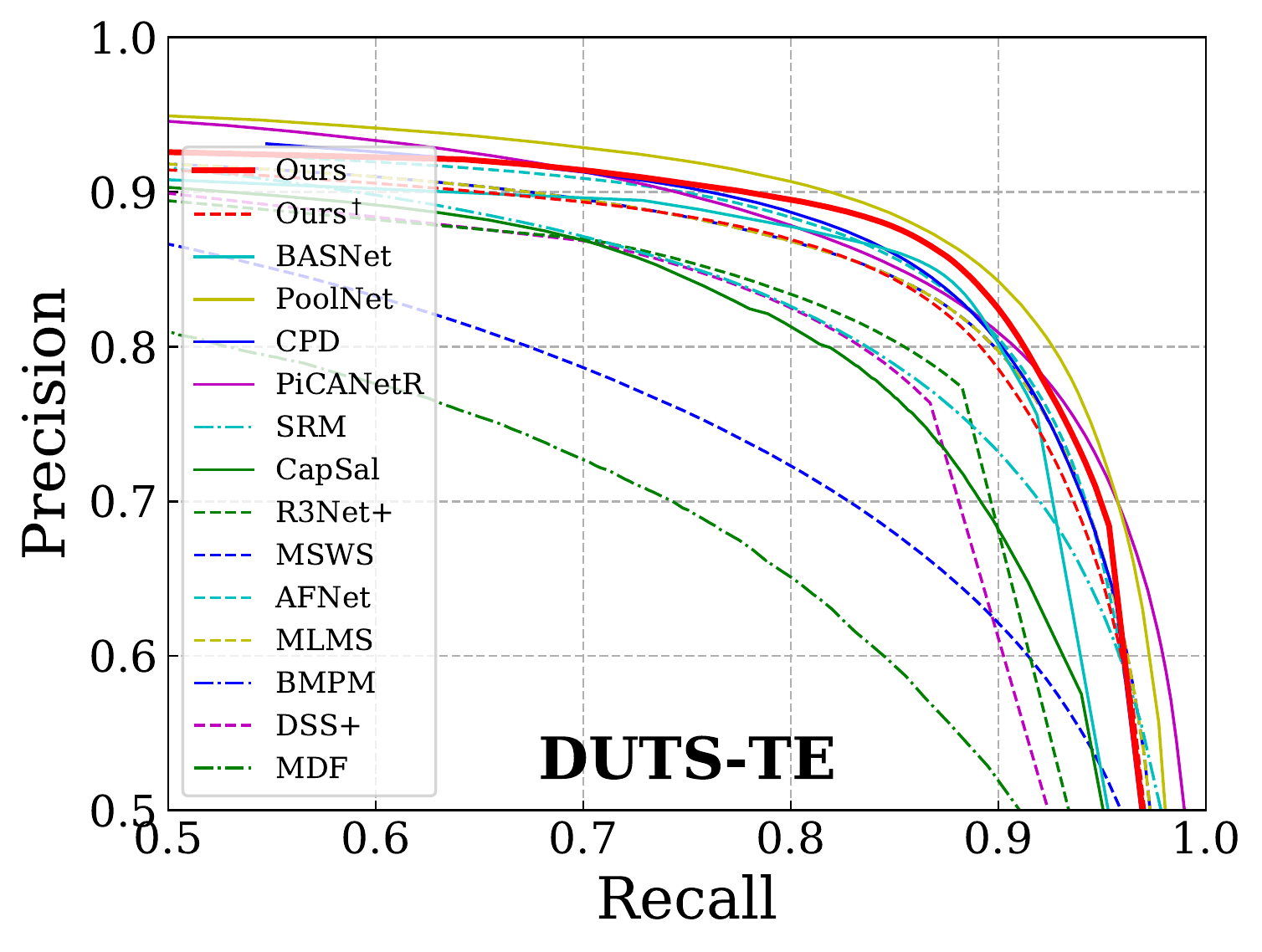}\\
    \includegraphics[width=0.49\linewidth]{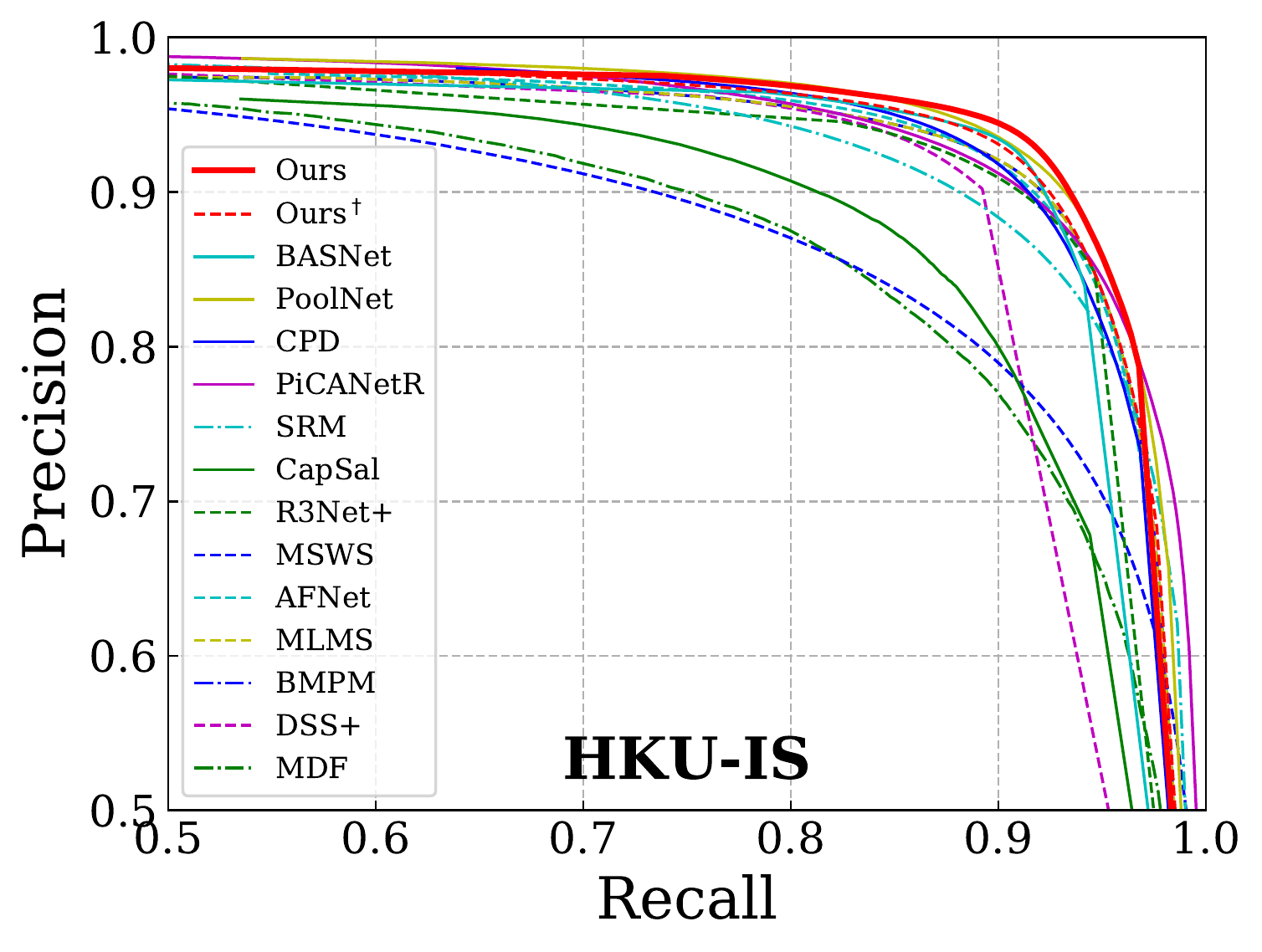}
    \vspace{-0.068in}
    \includegraphics[width=0.49\linewidth]{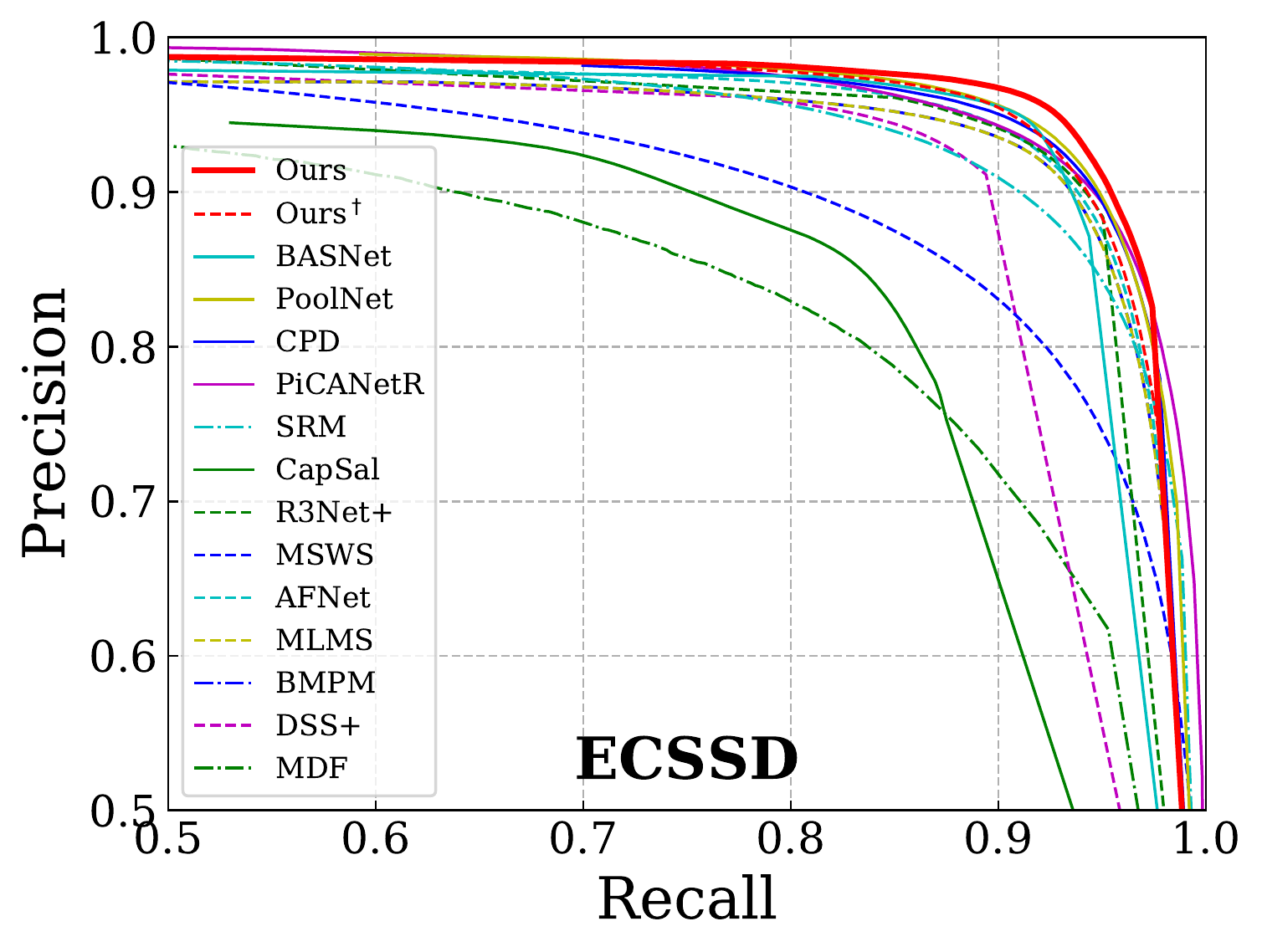}\\
    \includegraphics[width=0.49\linewidth]{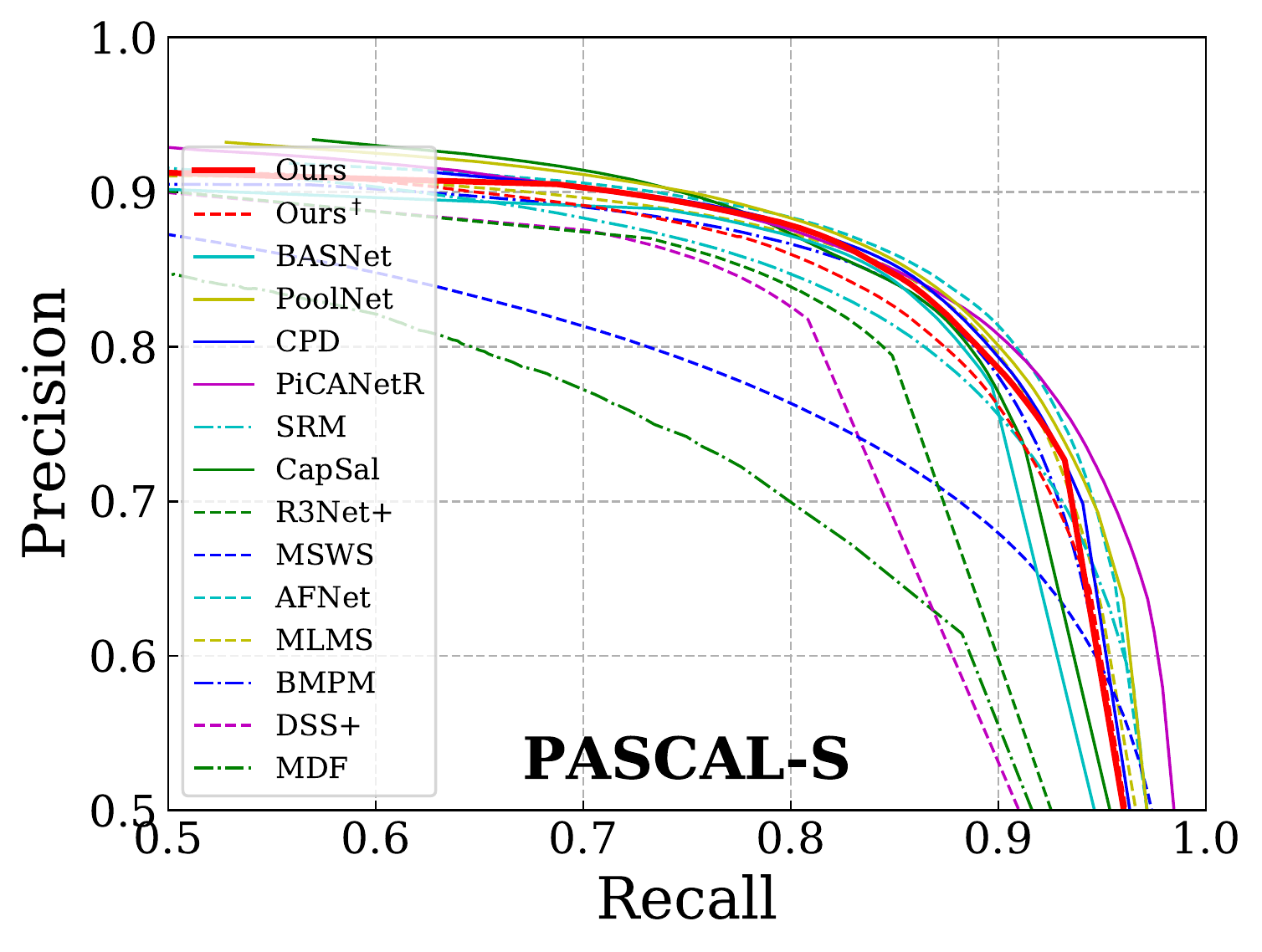}
    \includegraphics[width=0.49\linewidth]{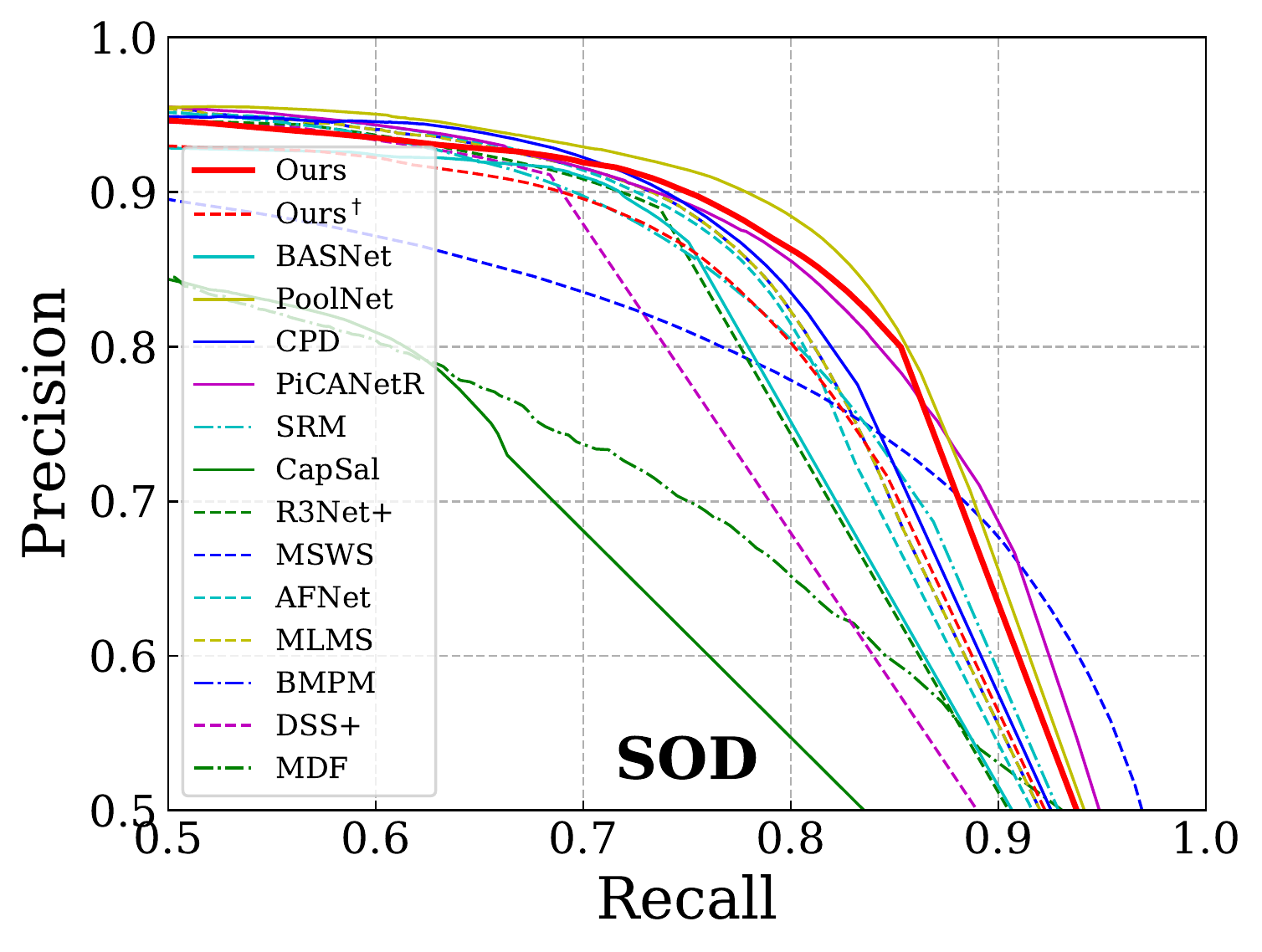}
    \centering
    \caption{Precision-Recall curves of our models and other typical state-of-the-art models on six SOD datasets.} 
    \label{fig:pr_curves}
\end{figure*}




\begin{table*}[htbp]
	\caption{\small Comparison of our method and 20 SOTA methods on DUT-OMRON, DUTS-TE, HKU-IS in terms of model size, $\textit{max}F_\beta$ ($\uparrow$), $MAE$ ($\downarrow$), weighted  $F^w_\beta$ ($\uparrow$), structure measure $S_m$ ($\uparrow$) and relax boundary F-measure $relaxF^b_\beta$ ($\uparrow$). \rrr Red, \kkk \ggg Green, \kkk and \bbb Blue \kkk indicate the best, second best and third best performance.}
	\centering
	\resizebox{\textwidth}{!}{ 
		\begin{tabular}{rccccccccccccccccc}
			\hline
			\multirow{2}{*}{\textbf{Method}} & 
			\multirow{2}{*}{\textbf{Backbone}} &
			\multirow{2}{*}{\textbf{Size(MB)}} &
			\multicolumn{5}{c}{\textbf{DUT-OMRON} (5168)} & \multicolumn{5}{c}{\textbf{DUTS-TE} (5019)} & \multicolumn{5}{c}{\textbf{HKU-IS} (4447)}\\
			\cline{4-18}
			&  & & $\text{\textit{max}}F_\beta$ & $MAE$ & $F_\beta^w$ & $S_m$ & $relaxF^b_{\beta}$ & $\text{\textit{max}}F_\beta$ & $MAE$ & $F_\beta^w$ & $S_m$ & $relaxF^b_{\beta}$ & $\text{\textit{max}}F_\beta$ & $MAE$ & $F_\beta^w$ & $S_m$ & $relaxF^b_{\beta}$ \\
			\hline 
			\textbf{MDF}$_\text{TIP16}$  	&	  AlexNet  	&	112.1 	&	0.694 	&	0.142 	&	0.565 	&	0.721 	&	0.406 	&	0.729 	&	0.099 	&	0.543 	&	0.723 	&	0.447 	&	0.860 	&	0.129 	&	0.564 	&	0.810 	&	0.594 	\\
			\hline 
			\textbf{UCF}$_\text{ICCV17}$ 	&	 VGG-16 	&	117.9	&	0.730 	&	0.120 	&	0.573 	&	0.760 	&	0.480 	&	0.773 	&	0.112 	&	0.596 	&	0.777 	&	0.518 	&	0.888 	&	0.062 	&	0.779 	&	0.875 	&	0.679 	\\
			\textbf{Amulet}$_\text{ICCV17}$ 	&	 VGG-16 	&	132.6 	&	0.743 	&	0.098 	&	0.626 	&	0.781 	&	0.528 	&	0.778 	&	0.084 	&	0.658 	&	0.796 	&	0.568 	&	0.897 	&	0.051 	&	0.817 	&	0.886 	&	0.716 	\\
			\textbf{NLDF+}$_\text{CVPR17}$ 	&	 VGG-16 	&	428.0 	&	0.753 	&	0.080 	&	0.634 	&	0.770 	&	0.514 	&	0.813 	&	0.065 	&	0.710 	&	0.805 	&	0.591 	&	0.902 	&	0.048 	&	0.838 	&	0.879 	&	0.694 	\\
			\textbf{DSS+}$_\text{CVPR17}$ 	&	 VGG-16 	&	237.0 	&	0.781 	&	0.063 	&	0.697 	&	0.790 	&	0.559 	&	0.825 	&	0.056 	&	0.755 	&	0.812 	&	0.606 	&	0.916 	&	0.040 	&	0.867 	&	0.878 	&	0.706 	\\
			\textbf{RAS}$_\text{ECCV18}$ 	&	 VGG-16 	&	\bbb 81.0 \kkk	&	0.786 	&	0.062 	&	0.695 	&	0.814 	&	0.615 	&	0.831 	&	0.059 	&	0.740 	&	0.828 	&	0.656 	&	0.913 	&	0.045 	&	0.843 	&	0.887 	&	0.748 	\\
			\textbf{PAGRN}$_\text{CVPR18}$ 	&	 VGG-19 	&	-	&	0.771 	&	0.071 	&	0.622 	&	0.775 	&	0.582 	&	0.854 	&	0.055 	&	0.724 	&	0.825 	&	0.692 	&	0.918 	&	0.048 	&	0.820 	&	0.887 	&	0.762 	\\
			\textbf{BMPM}$_\text{CVPR18}$ 	&	 VGG-16 	&	-	&	0.774 	&	0.064 	&	0.681 	&	0.809 	&	0.612 	&	0.852 	&	0.048 	&	0.761 	&	0.851 	&	0.699 	&	0.921 	&	0.039 	&	0.859 	&	0.907 	&	0.773 	\\
			\textbf{PiCANet}$_\text{CVPR18}$ 	&	 VGG-16 	&	153.3 	&	0.794 	&	0.068 	&	0.691 	&	0.826 	&	0.643 	&	0.851 	&	0.054 	&	0.747 	&	0.851 	&	0.704 	&	0.921 	&	0.042 	&	0.847 	&	0.906 	&	0.784 	\\
			\textbf{MLMS}$_\text{CVPR19}$	&	VGG-16	&	263.0 	&	0.774 	&	0.064 	&	0.681 	&	0.809 	&	0.612 	&	0.852 	&	0.048 	&	0.761 	&	0.851 	&	0.699 	&	0.921 	&	0.039 	&	0.859 	&	0.907 	&	0.773 	\\
			\textbf{AFNet}$_\text{CVPR19}$	&	 VGG-16 	&	143.0 	&	0.797 	&	\bbb 0.057 \kkk	&	0.717 	&	0.826 	&	0.635 	&	0.862 	&	0.046 	&	0.785 	&	0.855 	&	0.714 	&	0.923 	&	0.036 	&	0.869 	&	0.905 	&	0.772 	\\
			\hline 
			\textbf{MSWS}$_\text{CVPR19}$	&	 Dense-169 	&	\ggg 48.6 \kkk	&	0.718 	&	0.109 	&	0.527 	&	0.756 	&	0.362 	&	0.767 	&	0.908 	&	0.586 	&	0.749 	&	0.376 	&	0.856 	&	0.084 	&	0.685 	&	0.818 	&	0.438 	\\
			\hline 
			\textbf{R}$^3$\textbf{Net+}$_\text{IJCAI18}$ 	&	 ResNeXt 	&	215.0 	&	0.795 	&	0.063 	&	0.728 	&	0.817 	&	0.599 	&	0.828 	&	0.058 	&	0.763 	&	0.817 	&	0.601 	&	0.915 	&	0.036 	&	0.877 	&	0.895 	&	0.740 	\\
			\hline 
			\textbf{CapSal}$_\text{CVPR19}$	&	ResNet-101	&	-	&	0.699 	&	0.101 	&	0.482 	&	0.674 	&	0.396 	&	0.823 	&	0.072 	&	0.691 	&	0.808 	&	0.605 	&	0.882 	&	0.062 	&	0.782 	&	0.850 	&	0.654 	\\
			\textbf{SRM}$_\text{ICCV17}$ 	&	 ResNet-50 	&	189.0 	&	0.769 	&	0.069 	&	0.658 	&	0.798 	&	0.523 	&	0.826 	&	0.058 	&	0.722 	&	0.824 	&	0.592 	&	0.906 	&	0.046 	&	0.835 	&	0.887 	&	0.680 	\\
			\textbf{DGRL}$_\text{CVPR18}$ 	&	 ResNet-50 	&	646.1 	&	0.779 	&	0.063 	&	0.697 	&	0.810 	&	0.584 	&	0.834 	&	0.051 	&	0.760 	&	0.836 	&	0.656 	&	0.913 	&	0.037 	&	0.865 	&	0.897 	&	0.744 	\\
			\textbf{PiCANetR}$_\text{CVPR18}$	&	 ResNet-50 	&	197.2 	&	0.803 	&	0.065 	&	0.695 	&	0.832 	&	0.632 	&	0.860 	&	0.050 	&	0.755 	& \bbb	0.859 	\kkk &	0.696 	&	0.918 	&	0.043 	&	0.840 	&	0.904 	&	0.765 	\\
			\textbf{CPD}$_\text{CVPR19}$	&	 ResNet-50 	&	183.0 	&	0.797 	&	\ggg 0.056 \kkk	&	0.719 	&	0.825 	&	0.655 	& \bbb	0.865 	\kkk & \ggg	0.043 	\kkk &	0.795 	&	0.858 	& \bbb	0.741 	\kkk &	0.925 	&	0.034 	&	0.875 	&	0.905 	&	0.795 	\\
			\textbf{PoolNet}$_\text{CVPR19}$	&	 ResNet-50 	&	273.3 	&	\bbb 0.808 \kkk	&	\ggg 0.056 \kkk	&	0.729 	& \bbb	0.836 	\kkk &	0.675 	& \rrr	0.880 	\kkk & \rrr	0.040 	\kkk & \rrr	0.807 	\kkk & \rrr	0.871 	\kkk & \rrr	0.765 	\kkk & \ggg	0.932 	\kkk & \bbb	0.033 	\kkk & \bbb	0.881 	\kkk & \rrr	0.917 	\kkk & \ggg	0.811 	\kkk \\
			\textbf{BASNet}$_\text{CVPR19}$ 	&	 ResNet-34 	&	348.5 	&	0.805 	&	\ggg 0.056 \kkk	& \ggg	0.751 	\kkk & \bbb	0.836 	\kkk & \ggg	0.694 	\kkk &	0.860 	&	0.047 	& \bbb	0.803 	\kkk &	0.853 	& \ggg	0.758 	\ggg & \bbb	0.928 	\kkk & \ggg	0.032 	\kkk & \ggg	0.889 	\kkk & \bbb	0.909 	\kkk & \bbb	0.807 	\bbb \\
			\hline 
			\textbf{U}$^2$\textbf{-Net} (Ours) 	&	 RSU 	&	176.3 	&	\rrr 0.823 \kkk	&	\rrr 0.054 \kkk	& \rrr	0.757 	\kkk & \rrr	0.847 	\kkk & \rrr	0.702 	\kkk & \ggg	0.873 	\kkk & \bbb	0.044 	\kkk & \ggg	0.804 	\kkk & \ggg	0.861 	\kkk & \rrr	0.765 	\kkk & \rrr	0.935 	\kkk & \rrr	0.031 	\kkk & \rrr	0.890 	\kkk & \ggg	0.916 	\kkk & \rrr	0.812 	\kkk \\
			\textbf{U}$^2$\textbf{-Net}$^\dagger$ (Ours) 	&	 RSU 	&	\rrr 4.7 \kkk	&	\ggg 0.813 \kkk	&	0.060 	& \bbb	0.731 	\kkk & \ggg	0.837 	\kkk & \bbb	0.676 	\kkk &	0.852 	&	0.054 	&	0.763 	&	0.847 	&	0.723 	& \bbb	0.928 	\kkk &	0.037 	&	0.867 	&	0.908 	&	0.794 	\\
			\hline
		\end{tabular}
	}
	\label{tab:compSOA01}
\end{table*}
\begin{table*}[htbp!]
	\caption{\small Comparison of our method and 20 SOTA methods on ECSSD, PASCAL-S, SOD in terms of model size, $\textit{max}F_\beta$ ($\uparrow$), $MAE$ ($\downarrow$), weighted  $F^w_\beta$ ($\uparrow$), structure measure $S_m$ ($\uparrow$) and relax boundary F-measure $relaxF^b_\beta$ ($\uparrow$). \rrr Red, \kkk \ggg Green, \kkk and \bbb Blue \kkk indicate the best, second best and third best performance.}
	\centering
	\resizebox{\textwidth}{!}{ 
		\begin{tabular}{cccccccccccccccccc}
			\hline
			\multirow{2}{*}{\textbf{Method}} & 
			\multirow{2}{*}{\textbf{Backbone}} &
			\multirow{2}{*}{\textbf{Size(MB)}} &
			\multicolumn{5}{c}{\textbf{ECSSD} (1000)} & \multicolumn{5}{c}{\textbf{PASCAL-S} (850)} & \multicolumn{5}{c}{\textbf{SOD} (300)}\\
			\cline{4-18}
			&  & & $\text{\textit{max}}F_\beta$ & $MAE$ & $F_\beta^w$ & $S_m$ & $relaxF^b_{\beta}$ & $\text{\textit{max}}F_\beta$ & $MAE$ & $F_\beta^w$ & $S_m$ & $relaxF^b_{\beta}$ & $\text{\textit{max}}F_\beta$ & $MAE$ & $F_\beta^w$ & $S_m$ & $relaxF^b_{\beta}$ \\
			\hline 
			\textbf{MDF}$_\text{TIP16}$  	&	  AlexNet  	&	112.1 	&	0.832 	&	0.105 	&	0.705 	&	0.776 	&	0.472 	&	0.759 	&	0.142 	&	0.589 	&	0.696 	&	0.343 	&	0.746 	&	0.192 	&	0.508 	&	0.643 	&	0.311 	\\
			\hline 
			\textbf{UCF}$_\text{ICCV17}$ 	&	 VGG-16 	&	117.9	&	0.903 	&	0.069 	&	0.806 	&	0.884 	&	0.669 	&	0.814 	&	0.115 	&	0.694 	&	0.805 	&	0.493 	&	0.808 	&	0.148 	&	0.675 	&	0.762 	&	0.471 	\\
			\textbf{Amulet}$_\text{ICCV17}$ 	&	 VGG-16 	&	132.6 	&	0.915 	&	0.059 	&	0.840 	&	0.894 	&	0.711 	&	0.828 	&	0.100 	&	0.734 	&	0.818 	&	0.541 	&	0.798 	&	0.144 	&	0.677 	&	0.753 	&	0.454 	\\
			\textbf{NLDF+}$_\text{CVPR17}$ 	&	 VGG-16 	&	428.0 	&	0.905 	&	0.063 	&	0.839 	&	0.897 	&	0.666 	&	0.822 	&	0.098 	&	0.737 	&	0.798 	&	0.495 	&	0.841 	&	0.125 	&	0.709 	&	0.755 	&	0.475 	\\
			\textbf{DSS+}$_\text{CVPR17}$ 	&	 VGG-16 	&	237.0 	&	0.921 	&	0.052 	&	0.872 	&	0.882 	&	0.696 	&	0.831 	&	0.093 	&	0.759 	&	0.798 	&	0.499 	&	0.846 	&	0.124 	&	0.710 	&	0.743 	&	0.444 	\\
			\textbf{RAS}$_\text{ECCV18}$ 	&	 VGG-16 	&	\bbb 81.0 \kkk	&	0.921 	&	0.056 	&	0.857 	&	0.893 	&	0.741 	&	0.829 	&	0.101 	&	0.736 	&	0.799 	&	0.560 	&	0.851 	&	0.124 	&	0.720 	&	0.764 	&	0.544 	\\
			\textbf{PAGRN}$_\text{CVPR18}$ 	&	 VGG-19 	&	-	&	0.927 	&	0.061 	&	0.834 	&	0.889 	&	0.747 	&	0.847 	&	0.090 	&	0.738 	&	0.822 	&	0.594 	&	 - 	&	 - 	&	-	&	 -	&	 - 	\\
			\textbf{BMPM}$_\text{CVPR18}$ 	&	 VGG-16 	&	-	&	0.928 	&	0.045 	&	0.871 	&	0.911 	&	0.770 	&	0.850 	&	0.074 	&	0.779 	&	0.845 	&	0.617 	&	0.856 	&	0.108 	&	0.726 	&	0.786 	&	0.562 	\\
			\textbf{PiCANet}$_\text{CVPR18}$ 	&	 VGG-16 	&	153.3 	&	0.931 	&	0.046 	&	0.865 	&	0.914 	&	0.784 	&	0.856 	&	0.078 	&	0.772 	& \bbb	0.848 	\kkk &	0.612 	&	0.854 	& \ggg	0.103 	\kkk &	0.722 	& \bbb	0.789 	\kkk &	0.572 	\\
			\textbf{MLMS}$_\text{CVPR19}$	&	VGG-16	&	263.0 	&	0.928 	&	0.045 	&	0.871 	&	0.911 	&	0.770 	&	0.855 	&	0.074 	&	0.779 	&	0.844 	&	0.620 	&	0.856 	&	0.108 	&	0.726 	&	0.786 	&	0.562 	\\
			\textbf{AFNet}$_\text{CVPR19}$	&	 VGG-16 	&	143.0 	&	0.935 	&	0.042 	&	0.887 	&	0.914 	&	0.776 	& \ggg	0.863 	\kkk & \rrr	0.070 	\kkk & \ggg	0.798 	\kkk & \ggg	0.849 	\kkk &	0.626 	&	0.856 	&	0.111 	&	0.723 	&	0.774 	&	 - 	\\
			\hline 
			\textbf{MSWS}$_\text{CVPR19}$	&	 Dense-169 	&	\ggg 48.6 \kkk	&	0.878 	&	0.096 	&	0.716 	&	0.828 	&	0.411 	&	0.786 	&	0.133 	&	0.614 	&	0.768 	&	0.289 	&	0.800 	&	0.167 	&	0.573 	&	0.700 	&	0.231 	\\
			\hline 
			\textbf{R}$^3$\textbf{Net+}$_\text{IJCAI18}$ 	&	 ResNeXt 	&	215.0 	&	0.934 	&	0.040 	&	0.902 	&	0.910 	&	0.759 	&	0.834 	&	0.092 	&	0.761 	&	0.807 	&	0.538 	&	0.850 	&	0.125 	& \bbb	0.735 	\kkk &	0.759 	&	0.431 	\\
			\hline 
			\textbf{CapSal}$_\text{CVPR19}$	&	ResNet-101	&	-	&	0.874 	&	0.077 	&	0.771 	&	0.826 	&	0.574 	& \bbb	0.861 	\kkk & \bbb	0.073 	\kkk &	0.786 	&	0.837 	&	0.527 	&	0.773 	&	0.148 	&	0.597 	&	0.695 	&	0.404 	\\
			\textbf{SRM}$_\text{ICCV17}$ 	&	 ResNet-50 	&	189.0 	&	0.917 	&	0.054 	&	0.853 	&	0.895 	&	0.672 	&	0.838 	&	0.084 	&	0.758 	&	0.834 	&	0.509 	&	0.843 	&	0.128 	&	0.670 	&	0.741 	&	0.392 	\\
			\textbf{DGRL}$_\text{CVPR18}$ 	&	 ResNet-50 	&	646.1 	&	0.925 	&	0.042 	&	0.883 	&	0.906 	&	0.753 	&	0.848 	&	0.074 	&	0.787 	&	0.839 	&	0.569 	&	0.848 	&	0.106 	&	0.731 	&	0.773 	&	0.502 	\\
			\textbf{PiCANetR}$_\text{CVPR18}$	&	 ResNet-50 	&	197.2 	&	0.935 	&	0.046 	&	0.867 	&	0.917 	&	0.775 	&	0.857 	&	0.076 	&	0.777 	& \rrr	0.854 	\kkk &	0.598 	&	0.856 	& \bbb	0.104 	\kkk &	0.724 	& \ggg	0.790 	\kkk &	0.528 	\\
			\textbf{CPD}$_\text{CVPR19}$	&	 ResNet-50 	&	183.0 	&	0.939 	& \ggg	0.037 	\kkk & \bbb	0.898 	\kkk & \bbb	0.918 	\kkk &	0.811 	& \bbb	0.861 	\kkk & \ggg	0.071 	\kkk & \rrr	0.800 	\kkk & \bbb	0.848 	\kkk &	0.639 	& \bbb	0.860 	\kkk &	0.112 	&	0.714 	&	0.767 	&	0.556 	\\
			\textbf{PoolNet}$_\text{CVPR19}$	&	 ResNet-50 	&	273.3 	& \ggg 	0.944 	\kkk & \bbb	0.039 	\kkk &	0.896 	& \ggg	0.921 	\kkk & \bbb	0.813 	\kkk & \rrr	0.865 	\kkk &	0.075 	& \ggg	0.798 	\kkk &	0.832 	& \bbb	0.644 	\kkk & \rrr	0.871 	\kkk & \rrr	0.102 	\kkk & \rrr	0.759 	\kkk & \rrr	0.797 	\kkk & \ggg	0.606 	\kkk \\
			\textbf{BASNet}$_\text{CVPR19}$ 	&	 ResNet-34 	&	348.5 	&	0.942 	& \ggg	0.037 	\kkk & \ggg	0.904 	\kkk &	0.916 	& \ggg	0.826 	\kkk &	0.856 	&	0.076 	& \ggg	0.798 	\kkk &	0.838 	& \rrr	0.660 	\kkk &	0.851 	&	0.113 	&	0.730 	&	0.769 	& \bbb	0.603 	\kkk \\
			\hline 
			\textbf{U}$^2$\textbf{-Net} (Ours) 	&	 RSU 	&	176.3 	& \rrr	0.951 	\kkk & \rrr	0.033 	\kkk & \rrr	0.910 	\kkk & \rrr	0.928 	\kkk & \rrr	0.836 	\kkk &	0.859 	&	0.074 	& \bbb	0.797 	\kkk &	0.844 	& \ggg	0.657 	\kkk & \ggg	0.861 	\kkk &	0.108 	& \ggg	0.748 	\kkk &	0.786 	& \rrr	0.613 	\kkk \\
			\textbf{U}$^2$\textbf{-Net}$^\dagger$ (Ours) 	&	 RSU 	&	\rrr 4.7 \kkk	& \bbb 	0.943 	\kkk &	0.041 	&	0.885 	& \bbb	0.918 	\kkk &	0.808 	&	0.849 	&	0.086 	&	0.768 	&	0.831 	&	0.627 	&	0.841 	&	0.124 	&	0.697 	&	0.759 	&	0.559 	\\
			
			\hline
		\end{tabular}
	}
	\label{tab:compSOA02}
\end{table*}

\subsection{Comparison with State-of-the-arts}

We compare our models (full size U$^2$-Net, 176.3 MB and small size U$^2$-Net$^{\dagger}$, 4.7 MB) with 20 state-of-the-art methods including 
one \textbf{AlexNet} based model: 
\textbf{MDF}; 
10 \textbf{VGG} based models: 
\textbf{UCF}, 
\textbf{Amulet}, 
\textbf{NLDF}, 
\textbf{DSS}, 
\textbf{RAS}, 
\textbf{PAGRN}, 
\textbf{BMPM}, 
\textbf{PiCANet},
\textbf{MLMS},
\textbf{AFNet}; 
one \textbf{DenseNet} based model \textbf{MSWS}; 
one \textbf{ResNeXt} based model: \textbf{R$^3$Net}; 
and seven \textbf{ResNet} based models: 
\textbf{CapSal}, 
\textbf{SRM}, 
\textbf{DGRL}, 
\textbf{PiCANetR},
\textbf{CPD},
\textbf{PoolNet},
\textbf{BASNet}. 
For fair comparison, we mainly use the salient object detection results provided by the authors. 
For the missing results on certain datasets of some methods, we run their released code with their trained models on their suggested environment settings.


\subsubsection{Quantitative Comparison} 

Fig. \ref{fig:pr_curves} illustrates the precision-recall curves of our models (U$^2$-Net, 176.3 MB and U$^2$-Net$^{\dagger}$, 4.7 MB) and typical state-of-the-art methods on the six datasets. 
The curves are consistent with the Table \ref{tab:compSOA01} and \ref{tab:compSOA02} which demonstrate the state-of-the-art performance of our U$^2$-Net on DUT-OMRON, HKU-IS and ECSSD and competitive performance on other datasets. 
Table~\ref{tab:compSOA01} and \ref{tab:compSOA02} compares five (six include the model size) evaluation metrics and the model size of our proposed method with others. 
As we can see, our U$^2$-Net achieves the best performance on datasets DUT-OMRON, HKU-IS and ECSSD in terms of almost all of the five evaluation metrics. On DUTS-TE dataset our U$^2$-Net achieves the second best overall performance, which is slightly inferior to PoolNet. 
On PASCAL-S, the performance of our U$^2$-Net is slightly inferior to AFNet, CPD and PoolNet. 
It is worth noting that U$^2$-Net achieves the second best performance in terms of the boundary quality evaluation metric $\text{\textit{relax}}F^b_\beta$. 
On SOD dataset, PoolNet performs the best and our U$^2$-Net is the second best in terms of the overall performance. 

\begin{figure*}[h!]
    \centering
    \begin{minipage}[b]{1\linewidth}
            \includegraphics[width=0.085\linewidth]{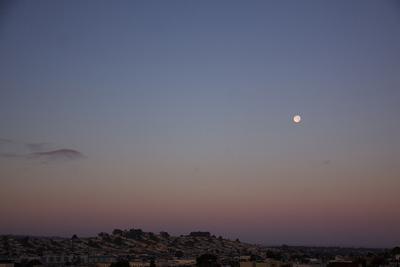}
            \includegraphics[width=0.085\linewidth]{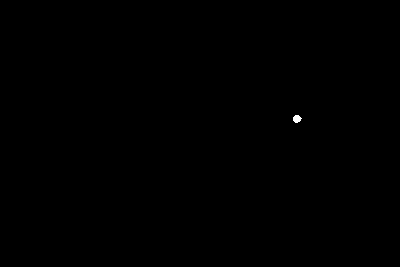}
            \includegraphics[width=0.085\linewidth]{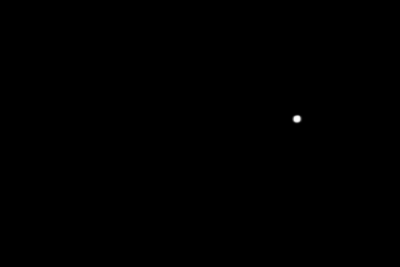}
            \includegraphics[width=0.085\linewidth]{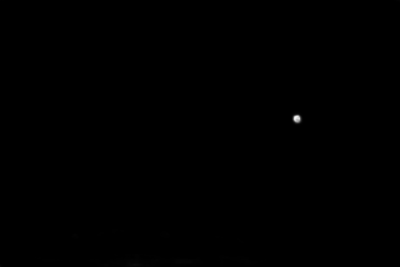}
            \includegraphics[width=0.085\linewidth]{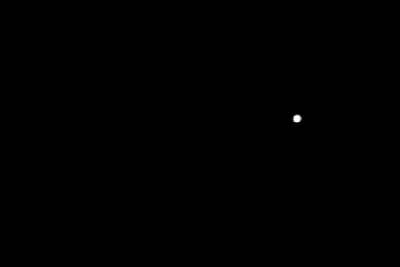}
            \includegraphics[width=0.085\linewidth]{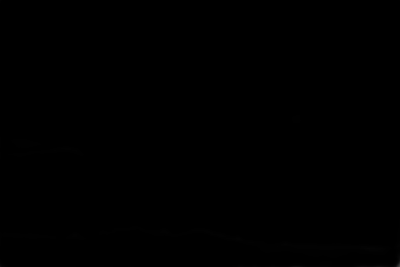}
            \includegraphics[width=0.085\linewidth]{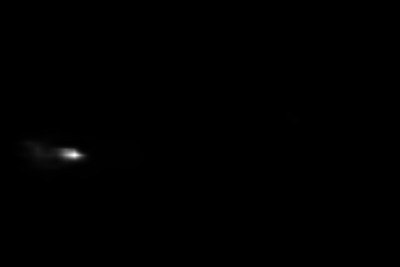}
            \includegraphics[width=0.085\linewidth]{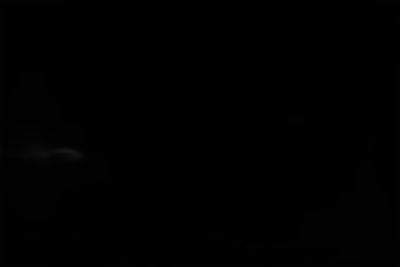}
            \includegraphics[width=0.085\linewidth]{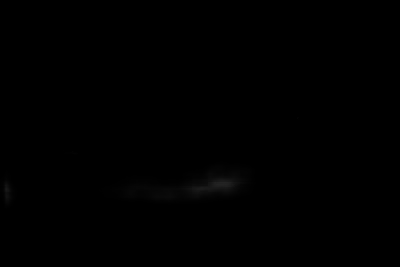}
            \includegraphics[width=0.085\linewidth]{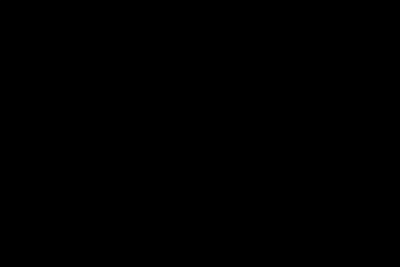}
            \includegraphics[width=0.085\linewidth]{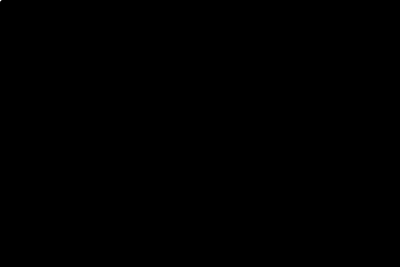}
        \end{minipage}\\
        \vspace{0.04in}
    \begin{minipage}[b]{1\linewidth}
            \includegraphics[width=0.085\linewidth]{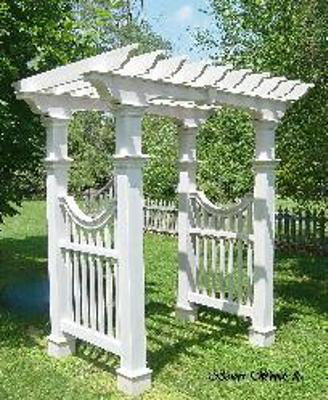}
            \includegraphics[width=0.085\linewidth]{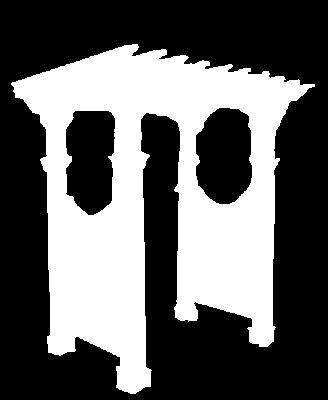}
            \includegraphics[width=0.085\linewidth]{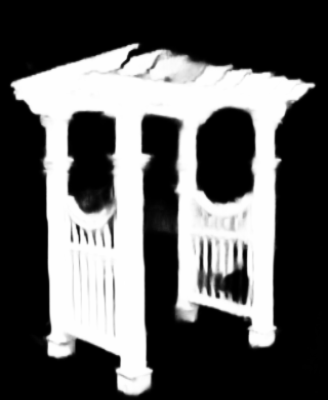}
            \includegraphics[width=0.085\linewidth]{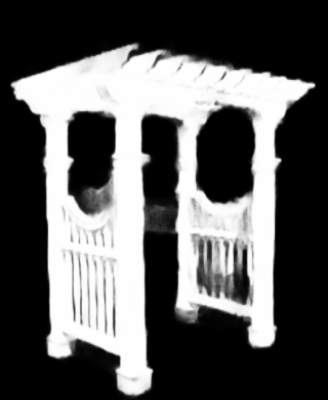}
            \includegraphics[width=0.085\linewidth]{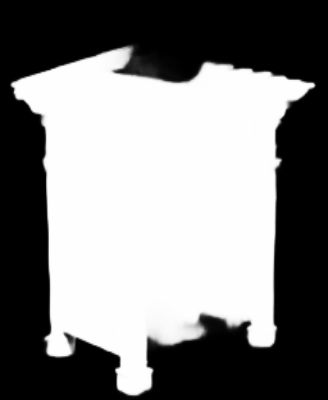}
            \includegraphics[width=0.085\linewidth]{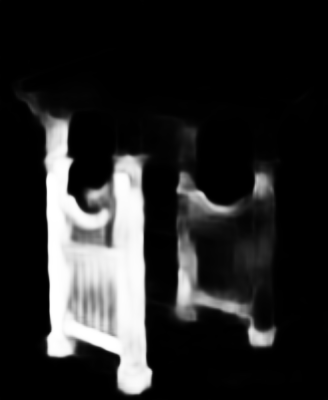}
            \includegraphics[width=0.085\linewidth]{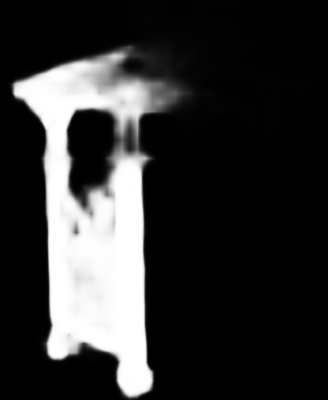}
            \includegraphics[width=0.085\linewidth]{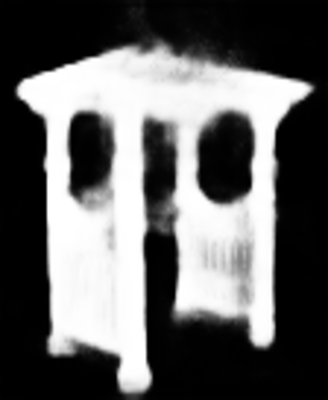}
            \includegraphics[width=0.085\linewidth]{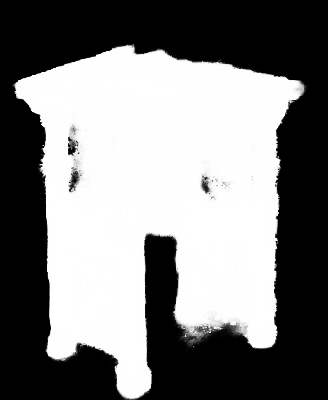}
            \includegraphics[width=0.085\linewidth]{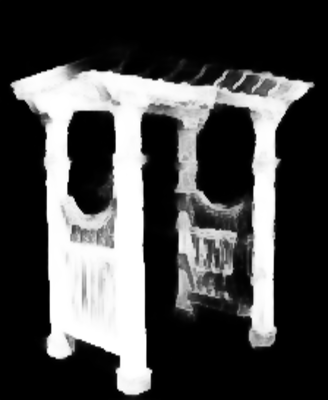}
            \includegraphics[width=0.085\linewidth]{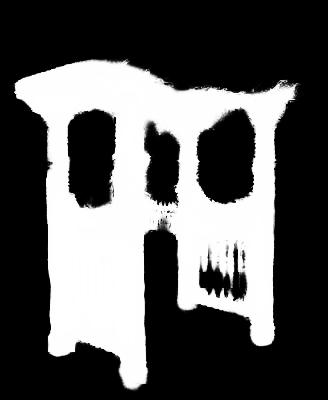}
        \end{minipage}\\
        \vspace{0.04in}
    \begin{minipage}[b]{1\linewidth}
            \includegraphics[width=0.085\linewidth]{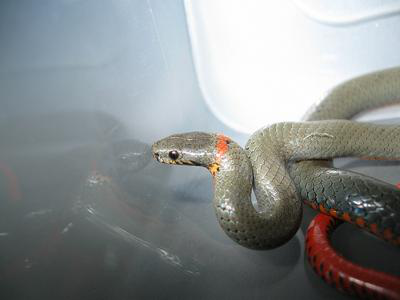}
            \includegraphics[width=0.085\linewidth]{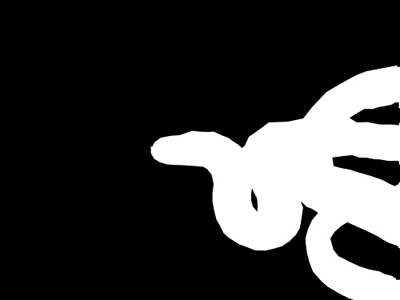}
            \includegraphics[width=0.085\linewidth]{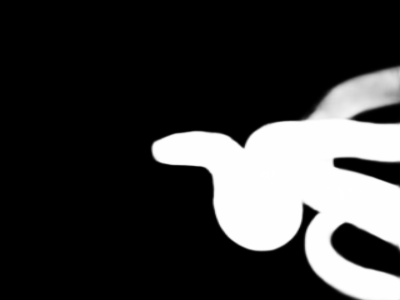}
            \includegraphics[width=0.085\linewidth]{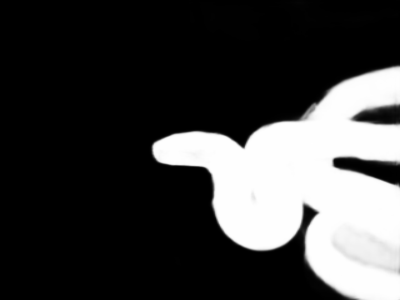}
            \includegraphics[width=0.085\linewidth]{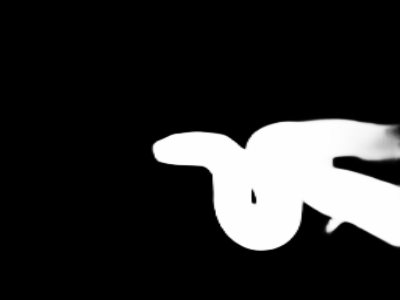}
            \includegraphics[width=0.085\linewidth]{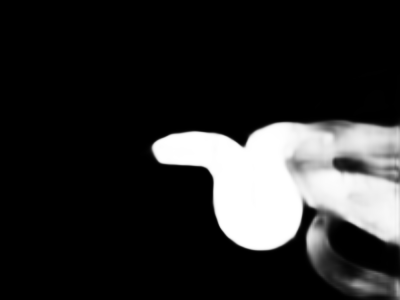}
            \includegraphics[width=0.085\linewidth]{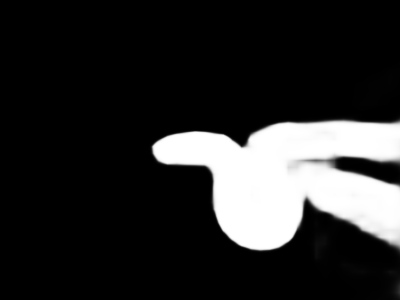}
            \includegraphics[width=0.085\linewidth]{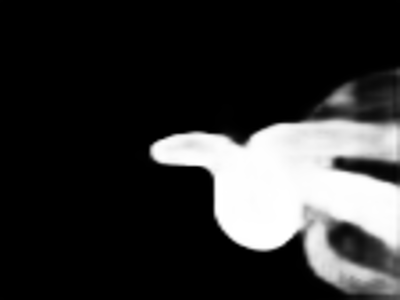}
            \includegraphics[width=0.085\linewidth]{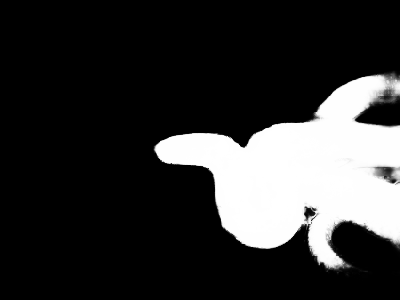}
            \includegraphics[width=0.085\linewidth]{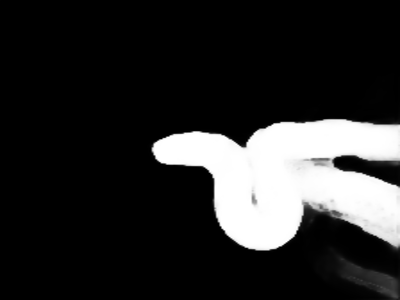}
            \includegraphics[width=0.085\linewidth]{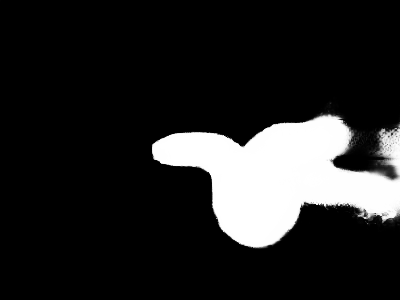}
        \end{minipage}\\
        \vspace{0.04in}
        \begin{minipage}[b]{1\linewidth}
            \includegraphics[width=0.085\linewidth]{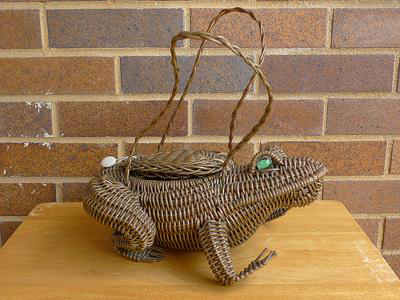}
            \includegraphics[width=0.085\linewidth]{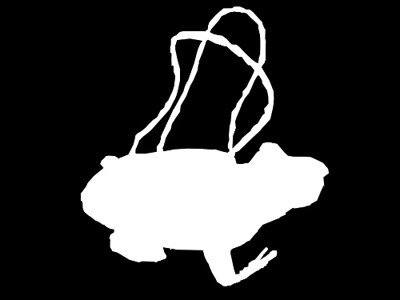}
            \includegraphics[width=0.085\linewidth]{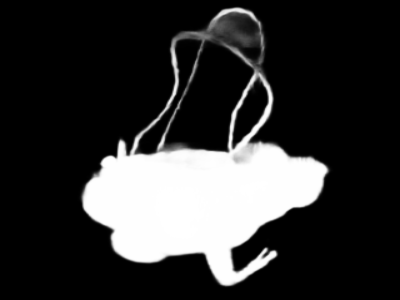}
            \includegraphics[width=0.085\linewidth]{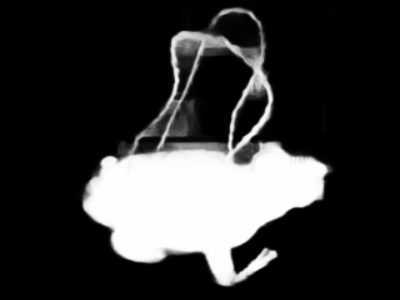}
            \includegraphics[width=0.085\linewidth]{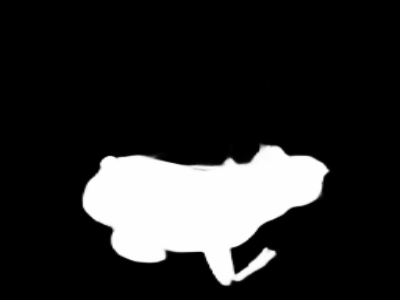}
            \includegraphics[width=0.085\linewidth]{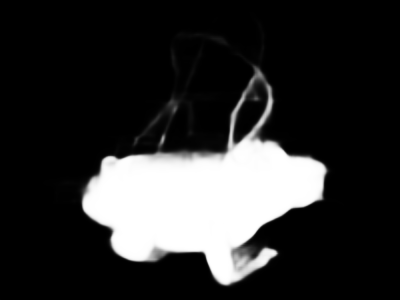}
            \includegraphics[width=0.085\linewidth]{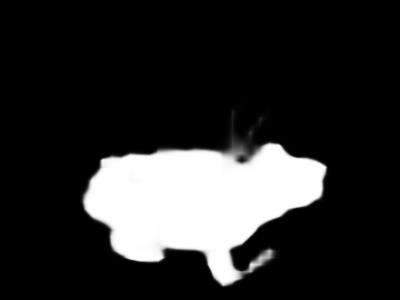}
            \includegraphics[width=0.085\linewidth]{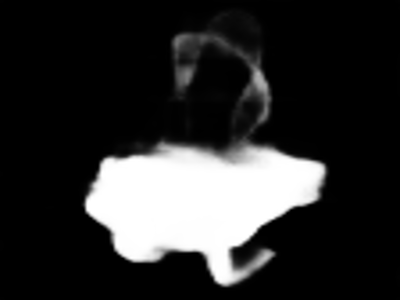}
            \includegraphics[width=0.085\linewidth]{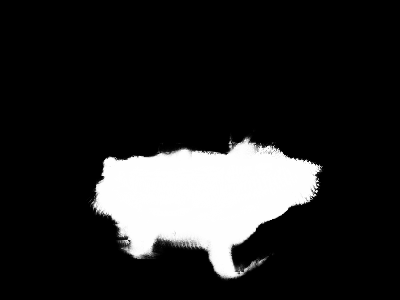}
            \includegraphics[width=0.085\linewidth]{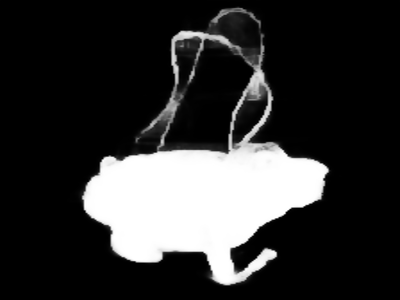}
            \includegraphics[width=0.085\linewidth]{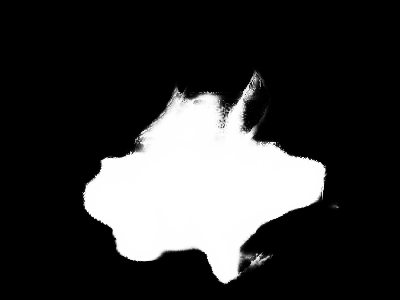}
        \end{minipage}\\
        \vspace{0.04in}
        \begin{minipage}[b]{1\linewidth}
            \includegraphics[width=0.085\linewidth]{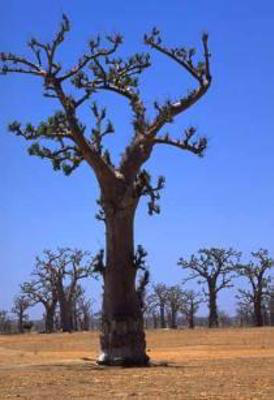}
            \includegraphics[width=0.085\linewidth]{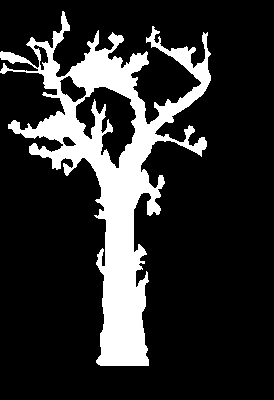}
            \includegraphics[width=0.085\linewidth]{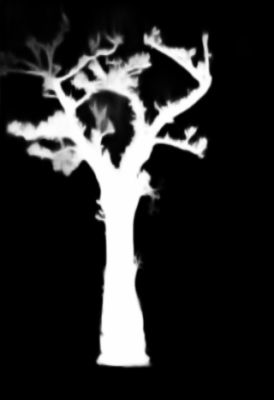}
            \includegraphics[width=0.085\linewidth]{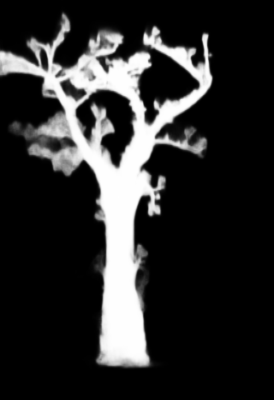}
            \includegraphics[width=0.085\linewidth]{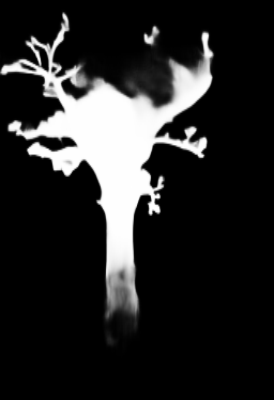}
            \includegraphics[width=0.085\linewidth]{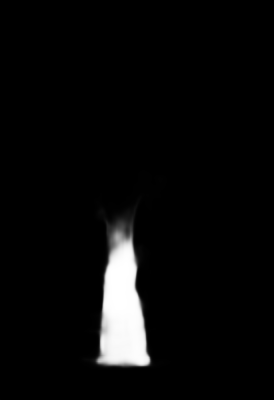}
            \includegraphics[width=0.085\linewidth]{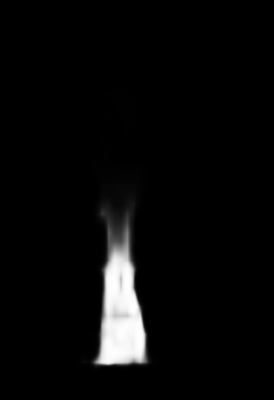}
            \includegraphics[width=0.085\linewidth]{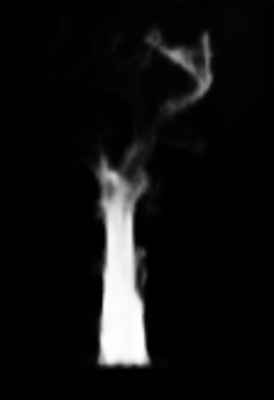}
            \includegraphics[width=0.085\linewidth]{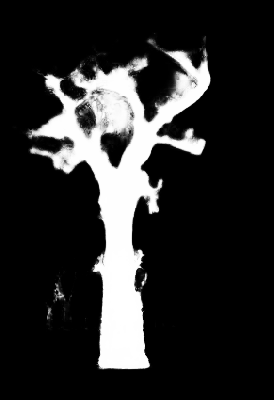}
            \includegraphics[width=0.085\linewidth]{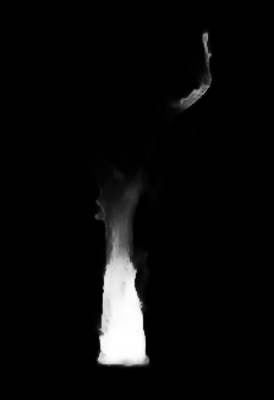}
            \includegraphics[width=0.085\linewidth]{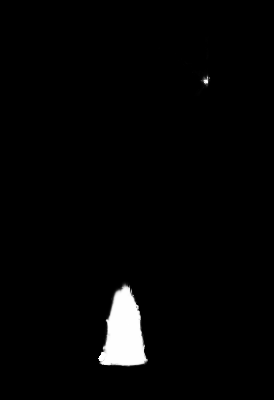}
        \end{minipage}\\
        \vspace{0.04in}
        \begin{minipage}[b]{1\linewidth}
            \vspace{-0.in}
            \includegraphics[width=0.085\linewidth]{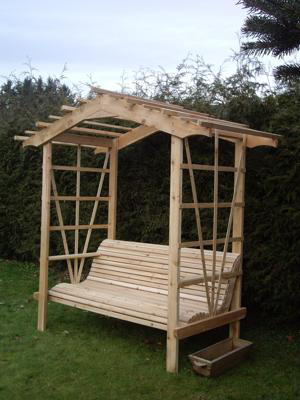}
            \includegraphics[width=0.085\linewidth]{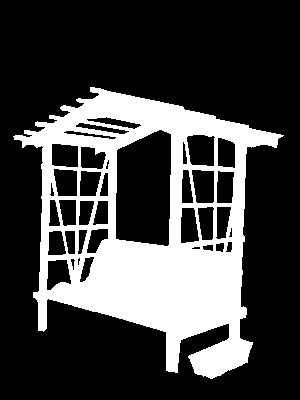}
            \includegraphics[width=0.085\linewidth]{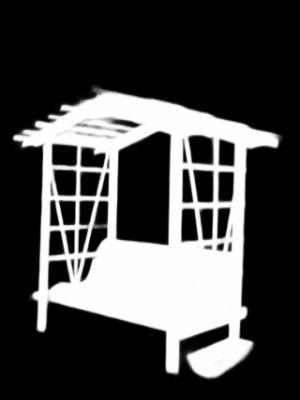}
            \includegraphics[width=0.085\linewidth]{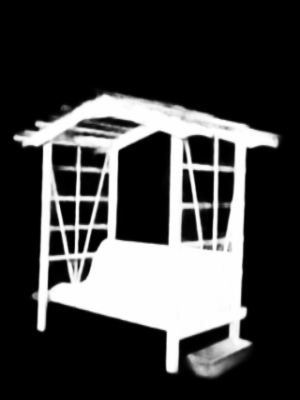}
            \includegraphics[width=0.085\linewidth]{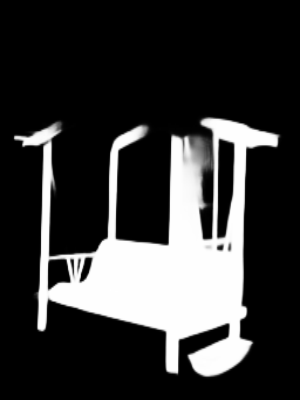}
            \includegraphics[width=0.085\linewidth]{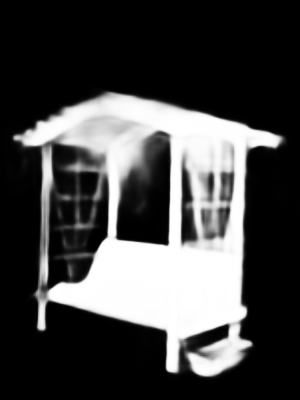}
            \includegraphics[width=0.085\linewidth]{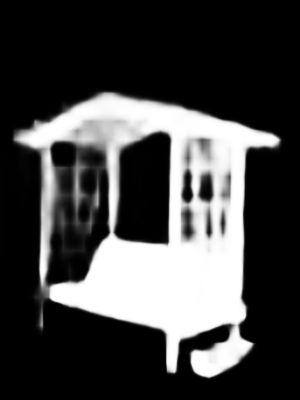}
            \includegraphics[width=0.085\linewidth]{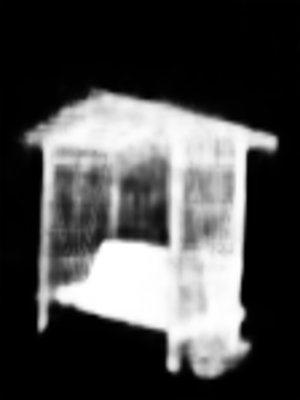}
            \includegraphics[width=0.085\linewidth]{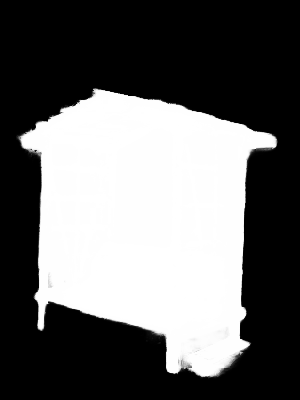}
            \includegraphics[width=0.085\linewidth]{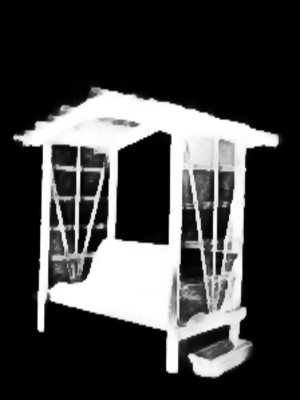}
            \includegraphics[width=0.085\linewidth]{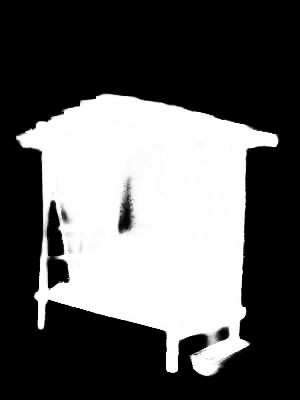}
        \end{minipage}\\
        \vspace{0.04in}
        \begin{minipage}[b]{1\linewidth}
            \vspace{-0.in}
            \includegraphics[width=0.085\linewidth]{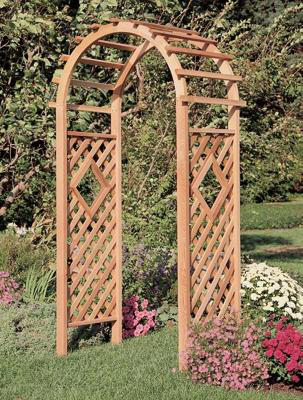}
            \includegraphics[width=0.085\linewidth]{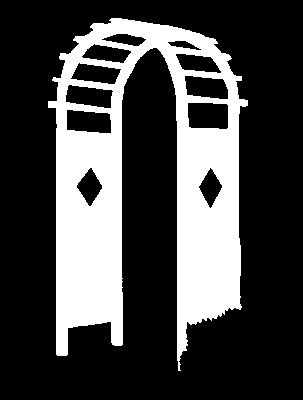}
            \includegraphics[width=0.085\linewidth]{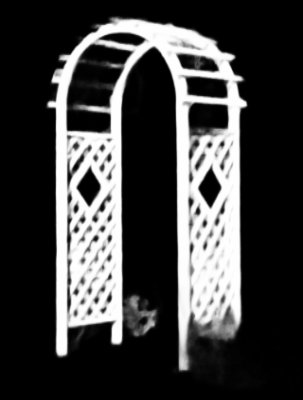}
            \includegraphics[width=0.085\linewidth]{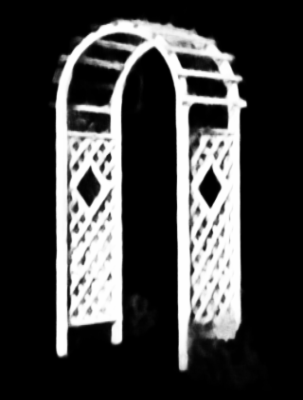}
            \includegraphics[width=0.085\linewidth]{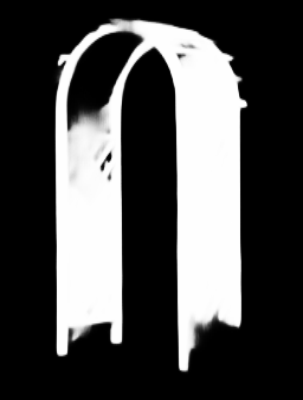}
            \includegraphics[width=0.085\linewidth]{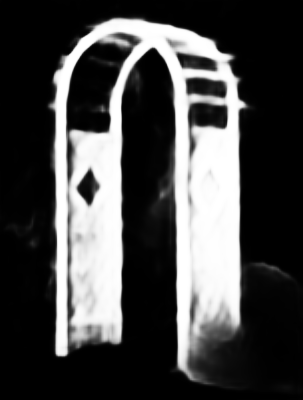}
            \includegraphics[width=0.085\linewidth]{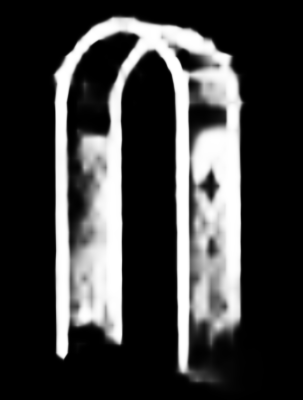}
            \includegraphics[width=0.085\linewidth]{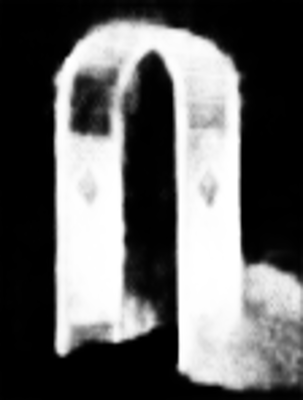}
            \includegraphics[width=0.085\linewidth]{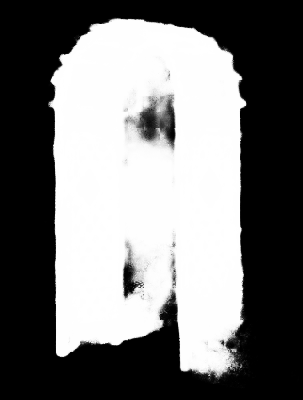}
            \includegraphics[width=0.085\linewidth]{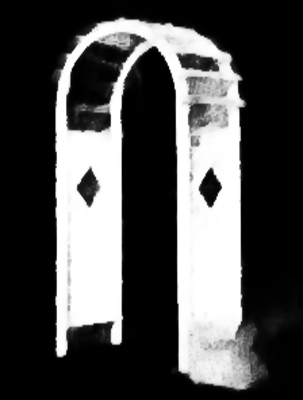}
            \includegraphics[width=0.085\linewidth]{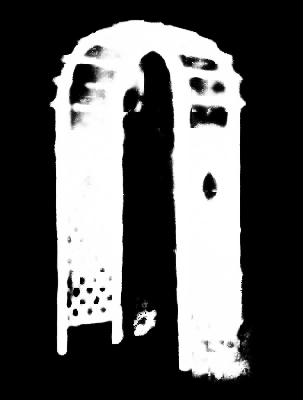}
        \end{minipage}\\
        \vspace{0.04in}
        \begin{minipage}[b]{1\linewidth}
            \vspace{-0.in}
            \includegraphics[width=0.085\linewidth]{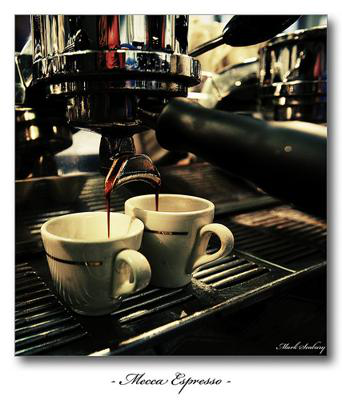}
            \includegraphics[width=0.085\linewidth]{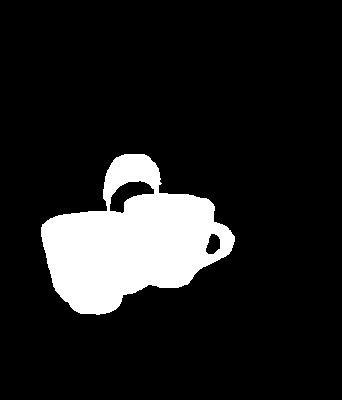}
            \includegraphics[width=0.085\linewidth]{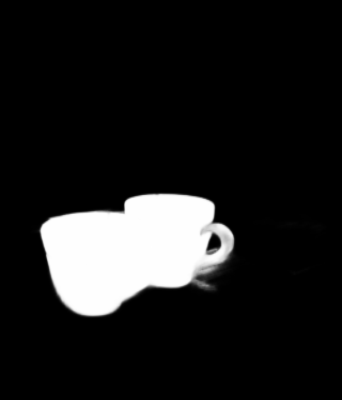}
            \includegraphics[width=0.085\linewidth]{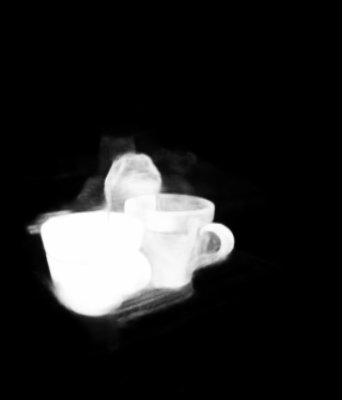}
            \includegraphics[width=0.085\linewidth]{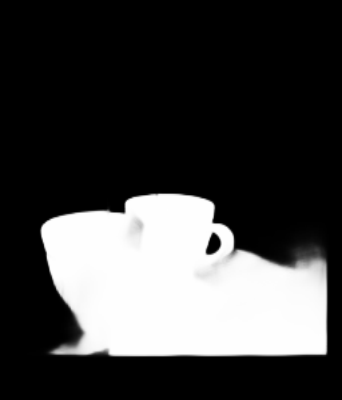}
            \includegraphics[width=0.085\linewidth]{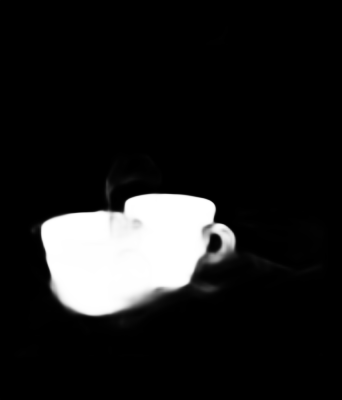}
            \includegraphics[width=0.085\linewidth]{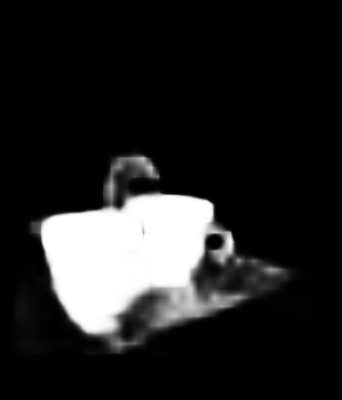}
            \includegraphics[width=0.085\linewidth]{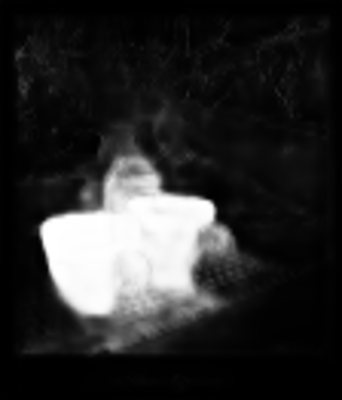}
            \includegraphics[width=0.085\linewidth]{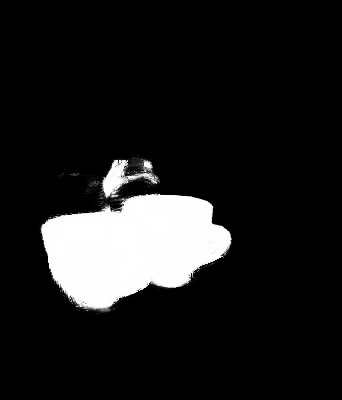}
            \includegraphics[width=0.085\linewidth]{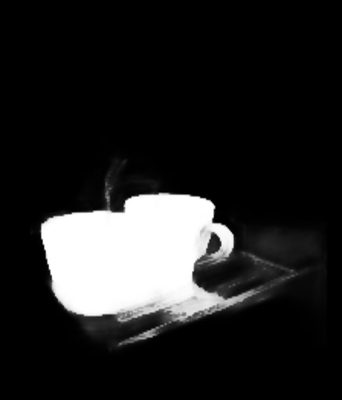}
            \includegraphics[width=0.085\linewidth]{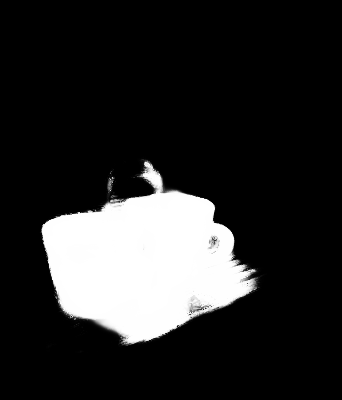}
        \end{minipage}\\
        \vspace{0.04in}
        \begin{minipage}[b]{1\linewidth}
            \vspace{-0.in}
            \includegraphics[width=0.085\linewidth]{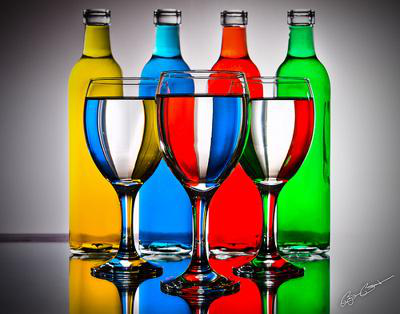}
            \includegraphics[width=0.085\linewidth]{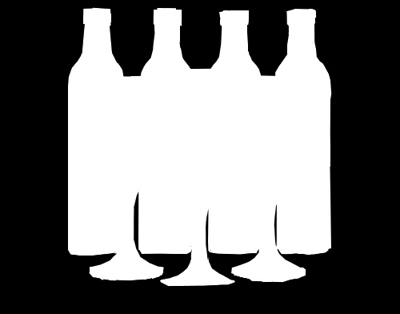}
            \includegraphics[width=0.085\linewidth]{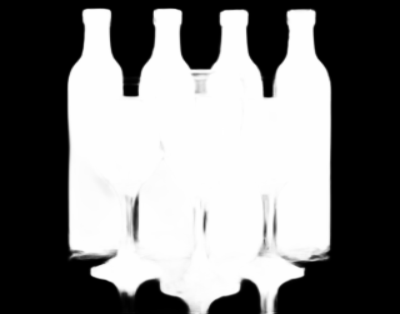}
            \includegraphics[width=0.085\linewidth]{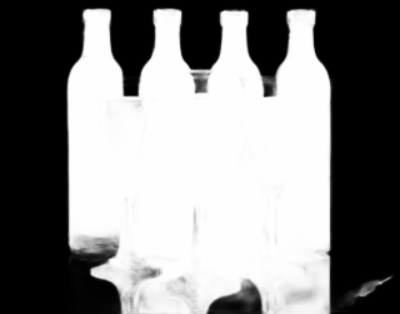}
            \includegraphics[width=0.085\linewidth]{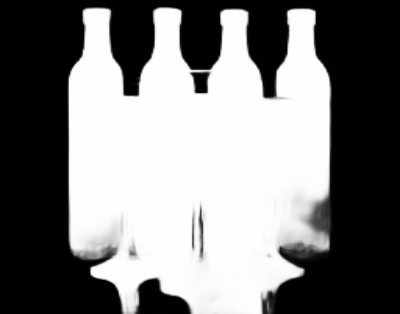}
            \includegraphics[width=0.085\linewidth]{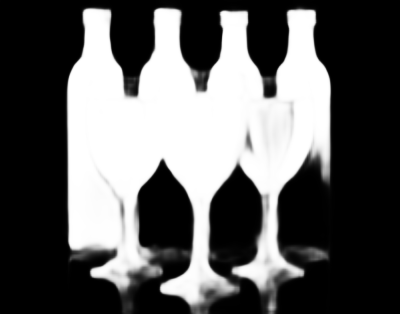}
            \includegraphics[width=0.085\linewidth]{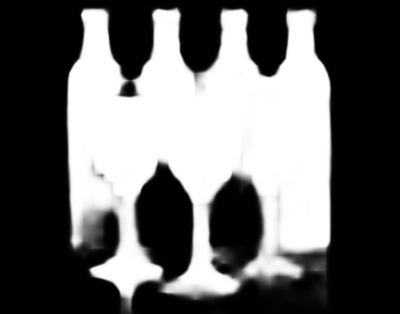}
            \includegraphics[width=0.085\linewidth]{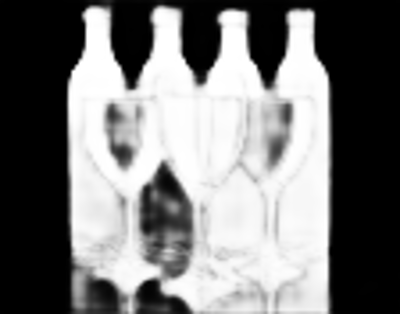}
            \includegraphics[width=0.085\linewidth]{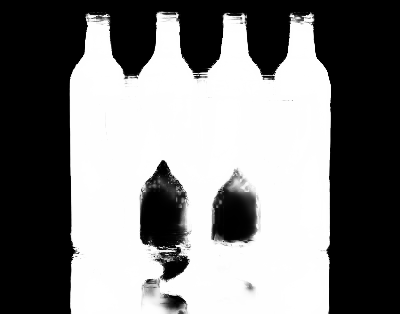}
            \includegraphics[width=0.085\linewidth]{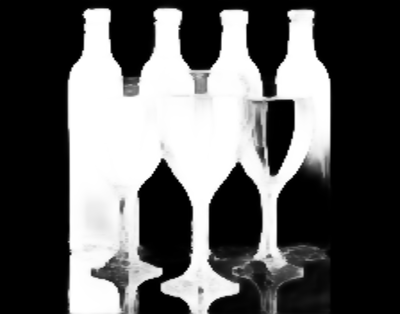}
            \includegraphics[width=0.085\linewidth]{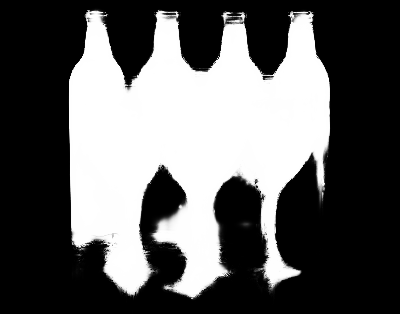}
        \end{minipage}\\
        \vspace{0.04in}
        \begin{minipage}[b]{1\linewidth}
            \vspace{-0.in}
            \subfigure[]{\includegraphics[width=0.085\linewidth]{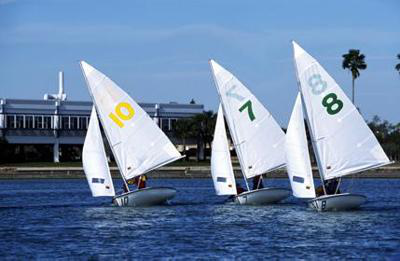}}
            \subfigure[]{\includegraphics[width=0.085\linewidth]{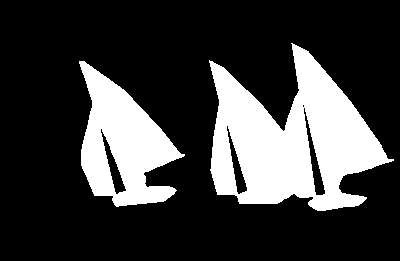}}
            \subfigure[]{\includegraphics[width=0.085\linewidth]{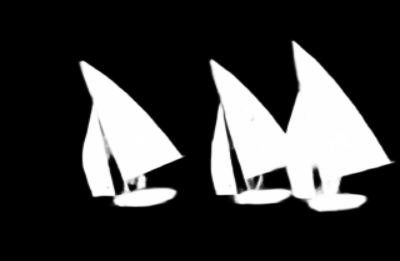}}
            \subfigure[]{\includegraphics[width=0.085\linewidth]{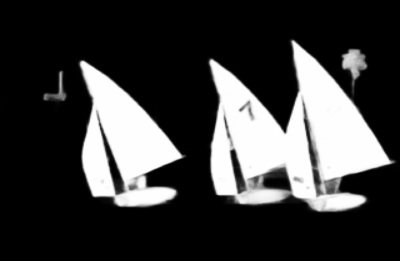}}
            \subfigure[]{\includegraphics[width=0.085\linewidth]{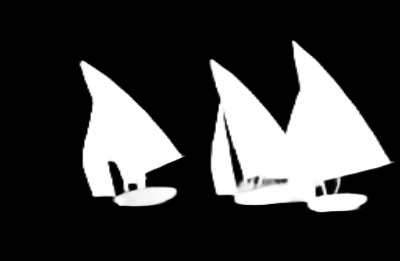}}
            \subfigure[]{\includegraphics[width=0.085\linewidth]{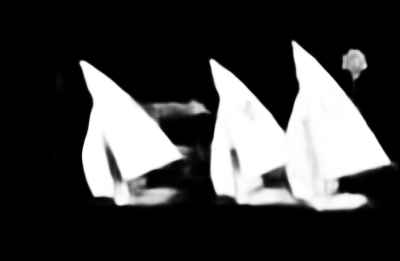}}
            \subfigure[]{\includegraphics[width=0.085\linewidth]{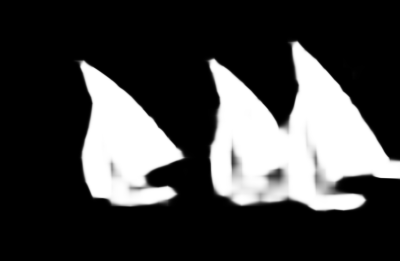}}
            \subfigure[]{\includegraphics[width=0.085\linewidth]{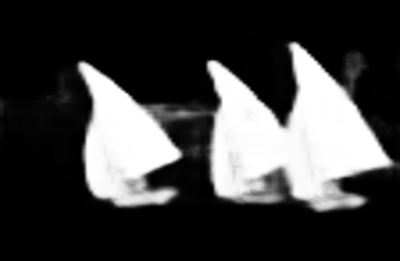}}
            \subfigure[]{\includegraphics[width=0.085\linewidth]{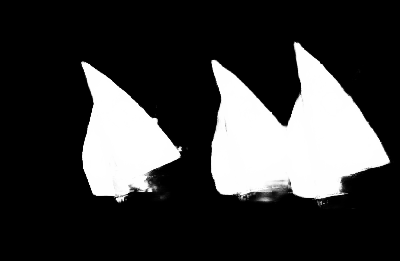}}
            \subfigure[]{\includegraphics[width=0.085\linewidth]{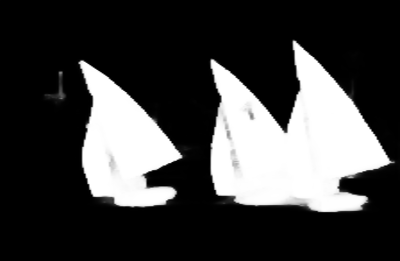}}
            \subfigure[]{\includegraphics[width=0.085\linewidth]{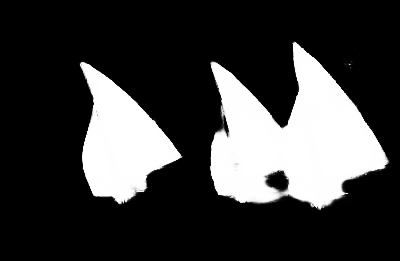}}
        \end{minipage}\\
    \caption{Qualitative comparison of the proposed method with seven other SOTA methods: (a) image, (b) GT, (c) Ours, (d) Ours$^{\dagger}$, (e) BASNet, (f) PoolNet, (g) CPD, (h) PiCANetR, (i) R$^3$Net+, (j) AFNet, (k) DSS+, where `+' indicates the CRF post-processing.}
    \label{fig:qual}
\end{figure*}

Our U$^2$-Net$^{\dagger}$ is only 4.7 MB, which is currently the smallest model in the field of salient object detection. 
With much fewer number of parameters against other models, it still achieves surprisingly competitive performance. 
Although its performance is not as good as our full size U$^2$-Net, its small size will facilitate its applications in many computation and memory constrained environments. 

\subsubsection{Qualitative Comparison:} 
To give an intuitive understanding of the promising performance of our models, we illustrate the sample results of our models and several other state-of-the-art methods in Fig.~\ref{fig:qual}. 
As we can see, our U$^2$-Net and U$^2$-Net$^{\dagger}$ are able to handle different types of targets and produce accurate salient object detection results. 

The 1st and 2nd row of Fig. \ref{fig:qual} show the results of small and large objects. As we can observe, our U$^2$-Net and U$^2$-Net$^\dagger$ are able to produce accurate results on both small and large objects. Other models either prone to miss the small target or produce large object with poor accuracy. 
The 3rd row shows the results of target touching image borders. 
Our U$^2$-Net correctly segments all the regions. Although U$^2$-Net$^\dagger$ erroneously segments the bottom right hole, it is still much better than other models. 
The 4th row demonstrates the performance of models in segmenting targets that consists of both large and thin structures. As we can see,  most of other models extract large regions well while missing the cable-wise thin structure except for AFNet (col (j)). 
The 5th row shows a tree with relatively clean background of blue sky. It seems easy, but it is actually challenging to most of the models because of the complicated shape of the target. As we can see, our models segment both the trunk and branches well, while others fail in segmenting the complicated tree branch region. 
Compared with the 5th row, the bench shown in the 6th row is more complex thanks to the hollow structure. Our U$^2$-Net produces near perfect result. Although the bottom right of the prediction map of U$^2$-Net$\dagger$ is imperfect, its overall performance on this target is much better than other models. Besides, the results of our models are more homogenous with fewer gray areas than models like PoolNet (col (f)), CPD (col (g)), PiCANetR (col (h)) and AFNet (col (j)). 
The 7th row shows that our models can produce results even finer than the ground truth. Labeling these small holes in the 7th image is burdensome and time-consuming. Hence, these repeated fine structures are usually ignored in the annotation process. Inferring the correct results from these imperfect labeling is challenging. But our models show promising capability in segmenting these fine structures thanks to the well designed architectures for extracting and integrating high resolution local and low resolution global information. The 8th and 9th row are illustrated to show the strong ability of our models in detecting targets with cluttered backgrounds and complicated foreground appearance.  The 10th row shows that our models are able to segment multiple targets while capturing the details of the detected targets (see the gap region of the two pieces of sail of each sailboat). 
In summary, both our full size and small size models are able to handle various scenarios and produce high accuracy salient object detection results.  

\section{Conclusions}

In this paper, we proposed a novel deep network: U$^2$-Net, for salient object detection. 
The main architecture of our U$^2$-Net is a two-level nested U-structure. 
The nested U-structure with our newly designed RSU blocks enables the network to capture richer local and global information from both shallow and deep layers regardless of the resolutions. 
Compared with those SOD models built upon the existing backbones, 
our U$^2$-Net is purely built on the proposed RSU blocks which makes it possible to be trained from scratch and configured to have different model size according to the target environment constraints. 
We provide a full size U$^2$-Net (176.3 MB, 30 FPS) and a smaller size version U$^2$-Net$^{\dagger}$ (4.7 MB, 40 FPS) in this paper. 
Experimental results on six public salient object detection datasets demonstrate that both models achieve very competitive performance against other 20 state-of-the-art methods in terms of both qualitative and quantitative measures. 

Although our models achieve competitive results against other state-of-the-art methods, faster and smaller models are needed for computation and memory limited devices, such as mobile phones, robots, etc. 
In the near future, we will explore different techniques and architectures to further improve the speed and decrease the model size.  
In addition, larger diversified salient object datasets are needed to train more accurate and robust models.

\section*{Acknowledgments}
	
This work is supported by the Alberta Innovates Graduate Student Scholarship and Natural Sciences and Engineering Research Council of Canada (NSERC) Discovery Grants Program, NO.: 2016-06365.

{\small
\bibliographystyle{ieee_fullname}
\bibliography{main.bib}

\begin{thebibliography}{10}\itemsep=-1pt

\bibitem{freqbased}
R. {Achanta}, S. {Hemami}, F. {Estrada}, and S. {Susstrunk}.
\newblock Frequency-tuned salient region detection.
\newblock In {\em 2009 IEEE Conference on Computer Vision and Pattern
  Recognition}, pages 1597--1604, 2009.

\bibitem{DBLP:journals/tip/BorjiCJL15}
Ali Borji, Ming{-}Ming Cheng, Huaizu Jiang, and Jia Li.
\newblock Salient object detection: {A} benchmark.
\newblock {\em {IEEE} Trans. Image Processing}, 24(12):5706--5722, 2015.

\bibitem{chen2017deeplab}
Liang-Chieh Chen, George Papandreou, Iasonas Kokkinos, Kevin Murphy, and Alan~L
  Yuille.
\newblock Deeplab: Semantic image segmentation with deep convolutional nets,
  atrous convolution, and fully connected crfs.
\newblock {\em IEEE transactions on pattern analysis and machine intelligence},
  40(4):834--848, 2017.

\bibitem{DBLP:conf/eccv/ChenTWH18}
Shuhan Chen, Xiuli Tan, Ben Wang, and Xuelong Hu.
\newblock Reverse attention for salient object detection.
\newblock In {\em Proceedings of the European Conference on Computer Vision
  (ECCV)}, pages 234--250, 2018.

\bibitem{imagenet_cvpr09}
Jia Deng, Wei Dong, Richard Socher, Li-Jia Li, Kai Li, and Li Fei-Fei.
\newblock Imagenet: A large-scale hierarchical image database.
\newblock In {\em 2009 IEEE conference on computer vision and pattern
  recognition}, pages 248--255. IEEE, 2009.

\bibitem{deng2018r3net}
Zijun Deng, Xiaowei Hu, Lei Zhu, Xuemiao Xu, Jing Qin, Guoqiang Han, and
  Pheng-Ann Heng.
\newblock R3net: Recurrent residual refinement network for saliency detection.
\newblock In {\em Proceedings of the 27th International Joint Conference on
  Artificial Intelligence}, pages 684--690. AAAI Press, 2018.

\bibitem{ehrig2005relaxed}
Marc Ehrig and J{\'e}r{\^o}me Euzenat.
\newblock Relaxed precision and recall for ontology matching.
\newblock In {\em Proc. K-Cap 2005 workshop on Integrating ontology}, pages
  25--32. No commercial editor., 2005.

\bibitem{fan2017structure}
Deng-Ping Fan, Ming-Ming Cheng, Yun Liu, Tao Li, and Ali Borji.
\newblock Structure-measure: A new way to evaluate foreground maps.
\newblock In {\em Proceedings of the IEEE Conference on Computer Vision and
  Pattern Recognition}, pages 4548--4557, 2017.

\bibitem{AFNet}
Mengyang Feng, Huchuan Lu, and Errui Ding.
\newblock Attentive feedback network for boundary-aware salient object
  detection.
\newblock In {\em Proceedings of the IEEE Conference on Computer Vision and
  Pattern Recognition}, pages 1623--1632, 2019.

\bibitem{DBLP:journals/jmlr/GlorotB10}
Xavier Glorot and Yoshua Bengio.
\newblock Understanding the difficulty of training deep feedforward neural
  networks.
\newblock In {\em Proceedings of the Thirteenth International Conference on
  Artificial Intelligence and Statistics, {AISTATS}}, pages 249--256, 2010.

\bibitem{haralick1987image}
Robert~M Haralick, Stanley~R Sternberg, and Xinhua Zhuang.
\newblock Image analysis using mathematical morphology.
\newblock {\em IEEE transactions on pattern analysis and machine intelligence},
  (4):532--550, 1987.

\bibitem{he2016deep}
Kaiming He, Xiangyu Zhang, Shaoqing Ren, and Jian Sun.
\newblock Deep residual learning for image recognition.
\newblock In {\em Proceedings of the IEEE conference on computer vision and
  pattern recognition}, pages 770--778, 2016.

\bibitem{hou2017deeply}
Qibin Hou, Ming-Ming Cheng, Xiaowei Hu, Ali Borji, Zhuowen Tu, and Philip Torr.
\newblock Deeply supervised salient object detection with short connections.
\newblock In {\em Proceedings of the IEEE Conference on Computer Vision and
  Pattern Recognition}, pages 5300--5309, 2017.

\bibitem{DBLP:conf/aaai/HuZQFH18}
Xiaowei Hu, Lei Zhu, Jing Qin, Chi{-}Wing Fu, and Pheng{-}Ann Heng.
\newblock Recurrently aggregating deep features for salient object detection.
\newblock In {\em AAAI}, pages 6943--6950, 2018.

\bibitem{huang2017densely}
Gao Huang, Zhuang Liu, Laurens van~der Maaten, and Kilian~Q Weinberger.
\newblock Densely connected convolutional networks.
\newblock In {\em Proceedings of the IEEE Conference on Computer Vision and
  Pattern Recognition}, pages 2261--2269, 2017.

\bibitem{kingma2014adam}
Diederik~P Kingma and Jimmy Ba.
\newblock Adam: A method for stochastic optimization.
\newblock {\em arXiv preprint}, 2014.

\bibitem{krizhevsky2012imagenet}
Alex Krizhevsky, Ilya Sutskever, and Geoffrey~E Hinton.
\newblock Imagenet classification with deep convolutional neural networks.
\newblock In {\em Advances in neural information processing systems}, pages
  1097--1105, 2012.

\bibitem{li2016visual}
Guanbin Li and Yizhou Yu.
\newblock Visual saliency detection based on multiscale deep cnn features.
\newblock {\em IEEE Transactions on Image Processing}, 25(11):5012--5024, 2016.

\bibitem{li2014secrets}
Yin Li, Xiaodi Hou, Christof Koch, James~M Rehg, and Alan~L Yuille.
\newblock The secrets of salient object segmentation.
\newblock In {\em Proceedings of the IEEE Conference on Computer Vision and
  Pattern Recognition}, pages 280--287, 2014.

\bibitem{DBLP:journals/pr/LiangZTBW18}
Jie Liang, Jun Zhou, Lei Tong, Xiao Bai, and Bin Wang.
\newblock Material based salient object detection from hyperspectral images.
\newblock {\em Pattern Recognition}, 76:476--490, 2018.

\bibitem{liu2019auto}
Chenxi Liu, Liang-Chieh Chen, Florian Schroff, Hartwig Adam, Wei Hua, Alan~L
  Yuille, and Li Fei-Fei.
\newblock Auto-deeplab: Hierarchical neural architecture search for semantic
  image segmentation.
\newblock In {\em Proceedings of the IEEE Conference on Computer Vision and
  Pattern Recognition}, pages 82--92, 2019.

\bibitem{PoolNet}
Jiang-Jiang Liu, Qibin Hou, Ming-Ming Cheng, Jiashi Feng, and Jianmin Jiang.
\newblock A simple pooling-based design for real-time salient object detection.
\newblock In {\em Proceedings of the IEEE Conference on Computer Vision and
  Pattern Recognition}, pages 3917--3926, 2019.

\bibitem{liu2018picanet}
Nian Liu, Junwei Han, and Ming-Hsuan Yang.
\newblock Picanet: Learning pixel-wise contextual attention for saliency
  detection.
\newblock In {\em Proceedings of the IEEE Conference on Computer Vision and
  Pattern Recognition}, pages 3089--3098, 2018.

\bibitem{long2015fully}
Jonathan Long, Evan Shelhamer, and Trevor Darrell.
\newblock Fully convolutional networks for semantic segmentation.
\newblock In {\em Proceedings of the IEEE conference on computer vision and
  pattern recognition}, pages 3431--3440, 2015.

\bibitem{lu2012saliency}
Shijian Lu and Joo-Hwee Lim.
\newblock Saliency modeling from image histograms.
\newblock In {\em European Conference on Computer Vision}, pages 321--332.
  Springer, 2012.

\bibitem{lu2013robust}
Shijian Lu, Cheston Tan, and Joo-Hwee Lim.
\newblock Robust and efficient saliency modeling from image co-occurrence
  histograms.
\newblock {\em IEEE transactions on pattern analysis and machine intelligence},
  36(1):195--201, 2013.

\bibitem{luo2017non}
Zhiming Luo, Akshaya Mishra, Andrew Achkar, Justin Eichel, Shaozi Li, and
  Pierre-Marc Jodoin.
\newblock Non-local deep features for salient object detection.
\newblock In {\em Proceedings of the IEEE Conference on Computer Vision and
  Pattern Recognition}, pages 6593--6601, 2017.

\bibitem{Ma2018DocUNetDI}
Ke Ma, Zhixin Shu, Xue Bai, Jue Wang, and Dimitris Samaras.
\newblock Docunet: Document image unwarping via a stacked u-net.
\newblock In {\em CVPR}, pages 4700--4709, 2018.

\bibitem{Margolin2014HowTE}
Ran Margolin, Lihi Zelnik-Manor, and Ayellet Tal.
\newblock How to evaluate foreground maps.
\newblock {\em 2014 IEEE Conference on Computer Vision and Pattern
  Recognition}, pages 248--255, 2014.

\bibitem{movahedi2010design}
Vida Movahedi and James~H Elder.
\newblock Design and perceptual validation of performance measures for salient
  object segmentation.
\newblock In {\em 2010 IEEE Computer Society Conference on Computer Vision and
  Pattern Recognition-Workshops}, pages 49--56. IEEE, 2010.

\bibitem{newell2016stacked}
Alejandro Newell, Kaiyu Yang, and Jia Deng.
\newblock Stacked hourglass networks for human pose estimation.
\newblock In {\em European conference on computer vision}, pages 483--499.
  Springer, 2016.

\bibitem{paszke2017automatic}
Adam Paszke, Sam Gross, Soumith Chintala, Gregory Chanan, Edward Yang, Zachary
  DeVito, Zeming Lin, Alban Desmaison, Luca Antiga, and Adam Lerer.
\newblock Automatic differentiation in pytorch.
\newblock In {\em Autodiff workshop on NIPS}, 2017.

\bibitem{BASNet}
Xuebin Qin, Zichen Zhang, Chenyang Huang, Chao Gao, Masood Dehghan, and Martin
  Jagersand.
\newblock Basnet: Boundary-aware salient object detection.
\newblock In {\em Proceedings of the IEEE Conference on Computer Vision and
  Pattern Recognition}, pages 7479--7489, 2019.

\bibitem{ronneberger2015u}
Olaf Ronneberger, Philipp Fischer, and Thomas Brox.
\newblock U-net: Convolutional networks for biomedical image segmentation.
\newblock In {\em International Conference on Medical image computing and
  computer-assisted intervention}, pages 234--241. Springer, 2015.

\bibitem{simonyan2014very}
Karen Simonyan and Andrew Zisserman.
\newblock Very deep convolutional networks for large-scale image recognition.
\newblock {\em arXiv preprint arXiv:1409.1556}, 2014.

\bibitem{szegedy2015going}
Christian Szegedy, Wei Liu, Yangqing Jia, Pierre Sermanet, Scott Reed, Dragomir
  Anguelov, Dumitru Erhan, Vincent Vanhoucke, and Andrew Rabinovich.
\newblock Going deeper with convolutions.
\newblock In {\em Proceedings of the IEEE conference on computer vision and
  pattern recognition}, pages 1--9, 2015.

\bibitem{tang2018quantized}
Zhiqiang Tang, Xi Peng, Shijie Geng, Lingfei Wu, Shaoting Zhang, and Dimitris
  Metaxas.
\newblock Quantized densely connected u-nets for efficient landmark
  localization.
\newblock In {\em Proceedings of the European Conference on Computer Vision
  (ECCV)}, pages 339--354, 2018.

\bibitem{tang2018cu}
Zhiqiang Tang, Xi Peng, Shijie Geng, Yizhe Zhu, and Dimitris~N Metaxas.
\newblock Cu-net: coupled u-nets.
\newblock {\em arXiv preprint arXiv:1808.06521}, 2018.

\bibitem{wang2017learning}
Lijun Wang, Huchuan Lu, Yifan Wang, Mengyang Feng, Dong Wang, Baocai Yin, and
  Xiang Ruan.
\newblock Learning to detect salient objects with image-level supervision.
\newblock In {\em Proceedings of the IEEE Conference on Computer Vision and
  Pattern Recognition}, pages 136--145, 2017.

\bibitem{DBLP:conf/iccv/WangBZZL17}
Tiantian Wang, Ali Borji, Lihe Zhang, Pingping Zhang, and Huchuan Lu.
\newblock A stagewise refinement model for detecting salient objects in images.
\newblock In {\em Proceedings of the IEEE International Conference on Computer
  Vision}, pages 4039--4048, 2017.

\bibitem{wang2018detect}
Tiantian Wang, Lihe Zhang, Shuo Wang, Huchuan Lu, Gang Yang, Xiang Ruan, and
  Ali Borji.
\newblock Detect globally, refine locally: A novel approach to saliency
  detection.
\newblock In {\em Proceedings of the IEEE Conference on Computer Vision and
  Pattern Recognition}, pages 3127--3135, 2018.

\bibitem{MLMS}
Runmin Wu, Mengyang Feng, Wenlong Guan, Dong Wang, Huchuan Lu, and Errui Ding.
\newblock A mutual learning method for salient object detection with
  intertwined multi-supervision.
\newblock In {\em Proceedings of the IEEE Conference on Computer Vision and
  Pattern Recognition}, pages 8150--8159, 2019.

\bibitem{CPD}
Zhe Wu, Li Su, and Qingming Huang.
\newblock Cascaded partial decoder for fast and accurate salient object
  detection.
\newblock In {\em Proceedings of the IEEE Conference on Computer Vision and
  Pattern Recognition}, pages 3907--3916, 2019.

\bibitem{xie2017aggregated}
Saining Xie, Ross Girshick, Piotr Doll{\'a}r, Zhuowen Tu, and Kaiming He.
\newblock Aggregated residual transformations for deep neural networks.
\newblock In {\em Proceedings of the IEEE Conference on Computer Vision and
  Pattern Recognition}, pages 5987--5995, 2017.

\bibitem{xie2015holistically}
Saining Xie and Zhuowen Tu.
\newblock Holistically-nested edge detection.
\newblock In {\em Proceedings of the IEEE international conference on computer
  vision}, pages 1395--1403, 2015.

\bibitem{yan2013hierarchical}
Qiong Yan, Li Xu, Jianping Shi, and Jiaya Jia.
\newblock Hierarchical saliency detection.
\newblock In {\em Proceedings of the IEEE Conference on Computer Vision and
  Pattern Recognition}, pages 1155--1162, 2013.

\bibitem{yang2013saliency}
Chuan Yang, Lihe Zhang, Huchuan Lu, Xiang Ruan, and Ming-Hsuan Yang.
\newblock Saliency detection via graph-based manifold ranking.
\newblock In {\em Proceedings of the IEEE Conference on Computer Vision and
  Pattern Recognition}, pages 3166--3173, 2013.

\bibitem{MSWS}
Yu Zeng, Yunzhi Zhuge, Huchuan Lu, Lihe Zhang, Mingyang Qian, and Yizhou Yu.
\newblock Multi-source weak supervision for saliency detection.
\newblock In {\em Proceedings of the IEEE Conference on Computer Vision and
  Pattern Recognition}, pages 6074--6083, 2019.

\bibitem{DBLP:journals/pr/ZhangEWZY17}
Jinxia Zhang, Krista~A. Ehinger, Haikun Wei, Kanjian Zhang, and Jingyu Yang.
\newblock A novel graph-based optimization framework for salient object
  detection.
\newblock {\em Pattern Recognition}, 64:39--50, 2017.

\bibitem{zhang2018bi}
Lu Zhang, Ju Dai, Huchuan Lu, You He, and Gang Wang.
\newblock A bi-directional message passing model for salient object detection.
\newblock In {\em Proceedings of the IEEE Conference on Computer Vision and
  Pattern Recognition}, pages 1741--1750, 2018.

\bibitem{CapSal}
Lu Zhang, Jianming Zhang, Zhe Lin, Huchuan Lu, and You He.
\newblock Capsal: Leveraging captioning to boost semantics for salient object
  detection.
\newblock In {\em Proceedings of the IEEE Conference on Computer Vision and
  Pattern Recognition}, pages 6024--6033, 2019.

\bibitem{DBLP:conf/ijcai/ZhangLLS18}
Pingping Zhang, Wei Liu, Huchuan Lu, and Chunhua Shen.
\newblock Salient object detection by lossless feature reflection.
\newblock In {\em IJCAI}, pages 1149--1155, 2018.

\bibitem{amulet17}
Pingping Zhang, Dong Wang, Huchuan Lu, Hongyu Wang, and Xiang Ruan.
\newblock Amulet: Aggregating multi-level convolutional features for salient
  object detection.
\newblock In {\em Proceedings of the IEEE International Conference on Computer
  Vision}, pages 202--211, 2017.

\bibitem{DBLP:conf/iccv/ZhangWLWY17}
Pingping Zhang, Dong Wang, Huchuan Lu, Hongyu Wang, and Baocai Yin.
\newblock Learning uncertain convolutional features for accurate saliency
  detection.
\newblock In {\em Proceedings of the IEEE International Conference on Computer
  Vision}, pages 212--221, 2017.

\bibitem{DBLP:journals/pr/ZhangHLPSH19}
Qiang Zhang, Zhen Huo, Yi Liu, Yunhui Pan, Caifeng Shan, and Jungong Han.
\newblock Salient object detection employing a local tree-structured low-rank
  representation and foreground consistency.
\newblock {\em Pattern Recognition}, 92:119--134, 2019.

\bibitem{zhang2018progressive}
Xiaoning Zhang, Tiantian Wang, Jinqing Qi, Huchuan Lu, and Gang Wang.
\newblock Progressive attention guided recurrent network for salient object
  detection.
\newblock In {\em Proceedings of the IEEE Conference on Computer Vision and
  Pattern Recognition}, pages 714--722, 2018.

\bibitem{PSPNet}
Hengshuang Zhao, Jianping Shi, Xiaojuan Qi, Xiaogang Wang, and Jiaya Jia.
\newblock Pyramid scene parsing network.
\newblock In {\em Proceedings of the IEEE Conference on Computer Vision and
  Pattern Recognition}, pages 2881--2890, 2017.

\bibitem{zhuge2019deep}
Yunzhi Zhuge, Yu Zeng, and Huchuan Lu.
\newblock Deep embedding features for salient object detection.
\newblock In {\em Proceedings of the AAAI Conference on Artificial
  Intelligence}, volume~33, pages 9340--9347, 2019.

\end{thebibliography}
}

\end{document}